\documentclass[journal]{IEEEtran}

\usepackage{hyperref}
\usepackage{cite}

\usepackage{amsmath}
\usepackage[pdftex]{graphicx}
\usepackage{xcolor}
\usepackage{multirow}
\usepackage[caption=false,font=normalsize,labelfont=sf,textfont=sf]{subfig}
\definecolor{amber}{rgb}{1.0, 0.75, 0.0}

\begin{document}

\title{Incremental Learning for Semantic Segmentation of Large-Scale Remote Sensing Data}

\author{Onur Tasar,~\IEEEmembership{Student member,~IEEE,}
        Yuliya Tarabalka,~\IEEEmembership{Senior member,~IEEE,}
        Pierre Alliez
     
\thanks{O. Tasar, Y. Tarabalka, and P.~Alliez are with Universit{\'e} C{\^o}te d'Azur, TITANE team, INRIA, 06902 Sophia Antipolis, France. E-mail: onur.tasar@inria.fr}}

\markboth{}%
{Shell \MakeLowercase{\textit{et al.}}: Bare Demo of IEEEtran.cls for IEEE Journals}

\maketitle

\begin{abstract}
In spite of remarkable success of the convolutional neural networks on semantic segmentation, they suffer from catastrophic forgetting: a significant performance drop for the already learned classes when new classes are added on the data, having no annotations for the old classes. We propose an incremental learning methodology, enabling to learn segmenting new classes without hindering dense labeling abilities for the previous classes, although the entire previous data are not accessible. The key points of the proposed approach are adapting the network to learn new as well as old classes on the new training data, and allowing it to remember the previously learned information for the old classes. For adaptation, we keep a frozen copy of the previously trained network, which is used as a memory for the updated network in absence of annotations for the former classes. The updated network minimizes a loss function, which balances the discrepancy between outputs for the previous classes from the memory and updated networks, and the mis-classification rate between outputs for the new classes from the updated network and the new ground-truth.  For remembering, we either regularly feed samples from the stored, little fraction of the previous data or use the memory network, depending on whether the new data are collected from completely different geographic areas or from the same city. 

Our experimental results prove that it is possible to add new classes to the network, while maintaining its performance for the previous classes, despite the whole previous training data are not available.
\end{abstract}


\IEEEpeerreviewmaketitle

\section{Introduction}

\IEEEPARstart{R}{ecent} improvements in satellite sensors have enabled us to capture massive amount of remote sensing data with high spatial resolution, as well as rich spectral information. Generation of maps from such a huge amount of satellite images and updating them automatically have been long standing problems, as they are crucial for a wide range of applications in domains such as agriculture, navigation, environmental management, urban monitoring, and mapping. In this context, having a strong classification system, which performs a high-quality, pixel-wise, large-scale classification is the most essential step. 

The task of dense labeling or semantic segmentation consists in assigning a thematic label to every image pixel. In the last decade, with the great advances in deep neural networks, notably convolutional neural networks (CNNs), it has been possible to obtain accurate segmentations \cite{maggiori2016high, volpi2017dense, audebert2016semantic}. Among the CNN-based approaches, U-net architecture \cite{ronneberger2015u} has gained a particular attention due to its success in various segmentation problems in different domains (e.g,. medical imaging and remote sensing). This network architecture consists of a contracting path that captures the context and a symmetric expanding path, enabling accurate localization. In addition to traditional encoder-decoder layers, U-net architecture uses skip connections, which combine low level features with the high level ones in the expanding path to increase precision of localization. Variants of this network \cite{huang2018large, iglovikov2018ternausnet, iglovikov2018ternausnetv2}, (e.g., U-net, including VGG-11 \cite{simonyan2014very} encoder and corresponding decoder) have been applied to remote sensing images and have shown a remarkable performance. Although it has been shown that CNNs can generate fine-grained segmentations from remote sensing images, most of the works validate their methodology on a small dataset collected from only one city, where some part is used as training data and the rest is used for validation. Working only on this kind of dataset prevents researches from exploring how well their classifier generalizes to different areas in the world, since the training as well as the validation data come from the same distribution. The recently released Inria Aerial Image Labeling Dataset \cite{maggiori2017can} contains training and test images from completely different cities, but it provides annotations for only building and non-building classes.

\begin{figure*}
	\centering
	\includegraphics[width=\textwidth]{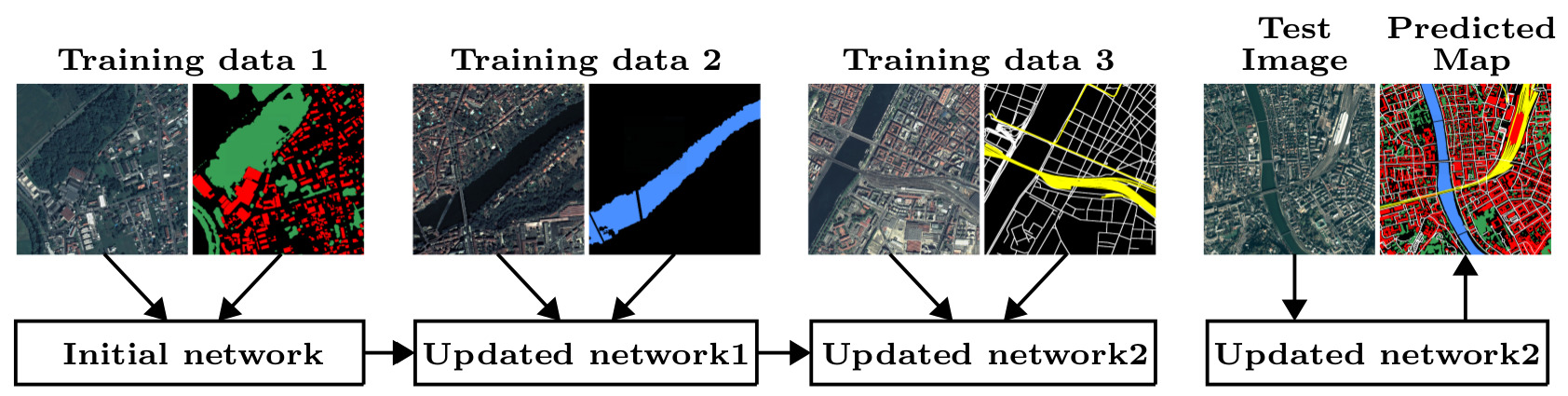}
  	\caption{An example incremental learning scenario. Firstly, satellite images as well as their label maps for building and high vegetation classes are fed to the network. Then, from the second training data, the network learns water class without forgetting building and high vegetation classes. Finally, road and railway classes are taught to the network. Whenever new training data are obtained, we store only a small part of the previous ones for the network to remember. When a new test image comes, the network is able to detect all the classes.}
	\label{fig:intro}
\end{figure*}

Another major drawback of the recently proposed methodologies is their assumption that the whole training data are available in the beginning, which is not the case in real world remote sensing applications, as new images are collected from all over the world everyday. Besides, having a large amount of standard, unique label map is almost impossible, because the label maps retrieved from different sources usually have distinct classes. In addition, it is not always possible to store enormous volume of training data. For the reasons described above, designing an incremental learning methodology, which can learn from the new training data while retaining performance for the old classes without accessing the entire previous training data is crucial. Although a good solution for this problem is necessary  to generate high-quality maps from satellite images that cover a large geographic extent, yet it has remained unexplored in remote sensing community.

Rather than assuming that we have all the training data initially, we aim to design an incremental learning methodology. Let us explain an example real-world problem (see Fig.~\ref{fig:intro}), where we are provided images as well as label maps for building and high vegetation classes from several cities in Austria in the beginning. Later on, we are given other training data, having label maps for water class, collected from different areas in Germany. Finally, we receive new satellite images and their annotations for road and railway classes from certain cities in France. Every time when the new data come, we assume that only a small portion of the previous data is stored. In such a scenario, our goal is to add segmentation capabilities for the new classes to the previously trained network without forgetting the already learned information so that maps for all the learned classes could be generated by the network. In addition to the described problem, because labeling satellite images, covering a large geographic area requires a lot of manual work, it is quite common that annotations of different classes for the same images are provided sequentially in time. In this kind of situation, whenever the new label maps are obtained, training a new classifier from scratch is not a feasible solution. The limitations pointed out in this section motivated us to design an incremental learning methodology.

\begin{figure*}
	\centering
	\includegraphics[width=1\linewidth]{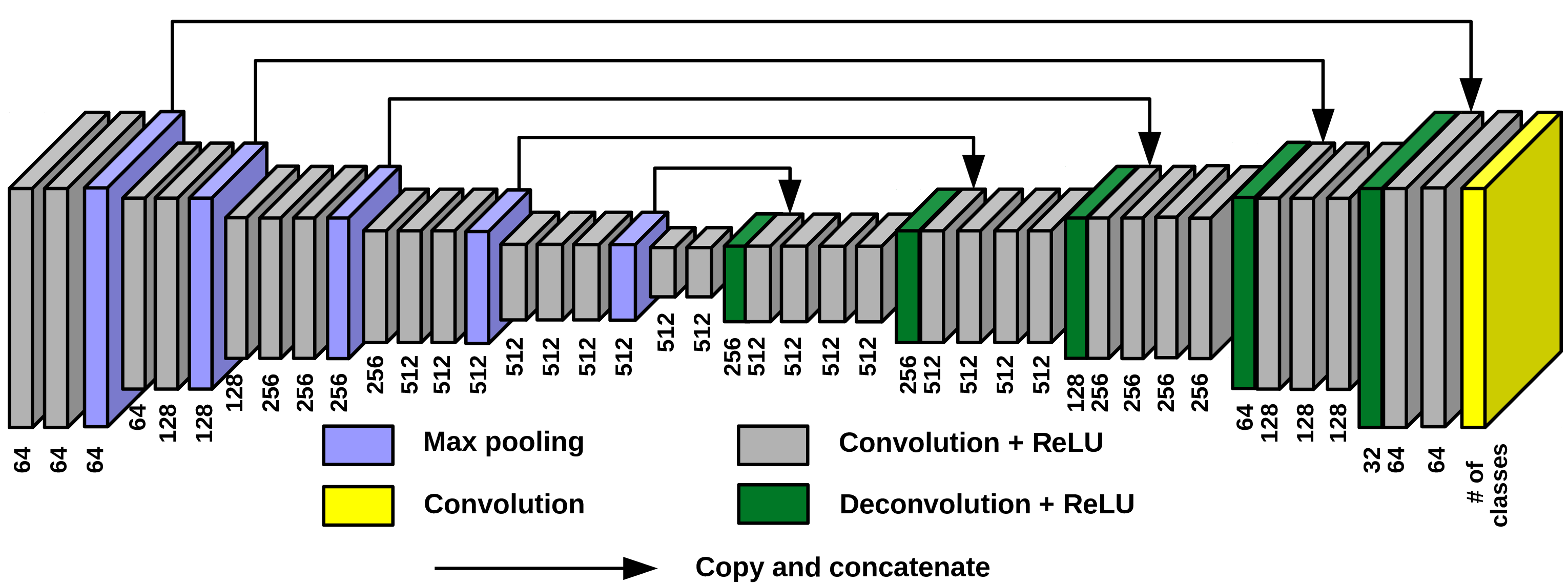}
	\caption{The network structure. The number below each layer corresponds to number of filters. We refer the last layer, shown by yellow color as the classification, and the rest as the shared layers.}
	\label{fig:network_structure}
\end{figure*}

\subsection{Related Work}
In this section, we summarize the proposed methodologies related to incremental learning problem.

The biggest challenge in incremental learning problem is that when the new tasks are intended to be added to a classification system, performance of the system for the previously learned tasks degrades abrubtly, which is referred as "catastrophic forgetting" in the literature \cite{french1999catastrophic, goodfellow2013empirical}. Incremental learning has been a historically important problem. Even before neural networks have become popular, researches had been studying this issue \cite{thrun1996learning, polikar2001learn++, cauwenberghs2001incremental, bruzzone1999incremental}. More recently, various convolutional neural network based methodologies have been proposed. There have been attempts, which change architecture of the neural network as the new classes are added. In \cite{sarwar2017incremental}, the network is trained incrementally by sharing early layers and splitting later ones by adding new convolutional kernels. In \cite{xiao2014error}, a tree-structured model, which grows hierarchically, is proposed. In \cite{rusu2016progressive} and \cite{yoon2018lifelong}, described approaches grow the network horizontally. The methodology described in \cite{xu2018reinforced} tries to solve the problem of determining number of filters to be added to each layer by reinforcement learning. The major weakness of these approaches is that since the network grows during training, the number of parameters increases drastically as the new tasks are added to the network. The methodologies proposed in \cite{rebuffi2017icarl, lopez2017gradient, hou2018lifelong} use not only the new training data but also a small portion of the old data. To determine the most important samples for the previous classes, the approach in \cite{li2018supportnet} trains a Support Vector Machine (SVM) from the previous training data. The support vectors of the SVM correspond to the samples to be used for the former classes, while the network is adapted to the new training data. In \cite{wu2018incremental, shin2017continual}, instead of using the old data directly, fake previous data are generated by generative adversarial networks (GANs). It has been proven that many configurations of the network parameters may produce the same result \cite{sussmann1992uniqueness}. Inspired by this idea, several works, which try to find a configuration of the network parameters that represents both the previous and the new training data well, have been published. The key idea behind these approaches is to find the important neurons for the old tasks and prevent these neurons from changing greatly or completely when the new tasks are added to the network. The proposed methodology explained in \cite{kirkpatrick2017overcoming} is one of the approaches that falls into this category. In the loss function defined in the paper, there is an elastic weight consolidation (EWC) term, which is multiplication of importance value of parameters for the old tasks and quadratic penalty on difference between parameters of the previous and the updated networks. Importance value of the parameters is measured by the estimated diagonal Fisher information matrix. The same work has been extended in \cite{liu2018rotate} by rotating the Fisher matrix. \cite{zenke2017continual} is also quite similar to \cite{kirkpatrick2017overcoming}, but the elastic weight consolidation is performed in online fashion. In \cite{aljundi2017memory}, importance of each neuron is determined by averaging gradients of the network output with respect to parameters of the neuron. In \cite{triki2017encoder}, in the training stage, features from the previous data are reconstructed in unsupervised manner using autoencoders. The features are then used to preserve information, which the old tasks rely on when the new tasks are added. \cite{lee2017overcoming} is another extension of \cite{kirkpatrick2017overcoming}, where trained models for all the tasks are combined via incremental moment matching (IMM). The proposed approaches in \cite{mallya2017packnet} and \cite{mallya2018piggyback} try to learn a mask, which marks important neurons for the old tasks. When the new tasks need to be added, only the masked out neurons are updated. In \cite{fernando2017pathnet}, paths through the network, which represent a subset of parameters are determined by using tournament selection genetic algorithm. During the training stage, only the neurons that are located along the paths are updated. When the data come sequentially, the works explained in \cite{zeno2018bayesian, ritter2018online, nguyen2017variational} optimize parameters of the network by updating the posterior approximation by the Bayesian inference based methods. Distilling the knowledge approach proposed in \cite{hinton2015distilling}, which enables to transfer the knowledge from a network or an assembly of several networks to a smaller network has inspired several works on incremental learning. The proposed methods in \cite{li2017lwf, furlanello2016active} facilitate a similar distillation loss described in \cite{hinton2015distilling} to maintain performance on the previous tasks. The proposed approach in \cite{castro2018end} uses a distillation loss function, which also uses samples for the previous classes in addition to samples for the new classes. Another knowledge distillation based approach has been proposed in \cite{shmelkov2017incremental}. They deal with incremental object detection and classification tasks at the same time. Although the incremental learning problem has been explored in depth in the literature, none of the works described in this section studies incremental learning for dense labeling.

\subsection{Contributions}
We propose a novel incremental learning methodology for semantic segmentation problem, where the network learns segmenting new classes without deteriorating performance for the previously learned classes, even when the entire previous training data are not stored\footnote{Project page: \url{https://project.inria.fr/epitome/inc_learn}}.

We deal with two common real-world problems, in which the former is the situation of retrieving stream of training data, where at each time step, the data contain satellite images collected from different locations in the world and annotations for separate classes, the latter is the case, where label maps for the same geographic area are provided sequentially. To investigate how our methodology performs on the first problem, we test our approach on the Luxcarta dataset, consisting of the satellite images captured over different cities in France and Austria. For the second problem, we conduct experiments on the Vaihingen and the Potsdam benchmark datasets provided by the ISPRS \cite{isprs_data}. The first problem is much more challenging, as the satellite images have high color variations and visual feature differences. Besides, for the first problem, by following a similar strategy described in \cite{maggiori2017can}, we test the trained models on the data collected from completely different geograhic areas than the ones we use during training. The benchmark dataset presented in \cite{maggiori2017can} enables to study how well the trained models generalize to new cities for only building class. Rather than doing a binary classification, we investigate how well the trained models perform on segmenting completely different geographic extents for multiple classes. We provide rich experimental results for both problems by comparing our methodology with the ones explained in Sec.~\ref{sec:compared_methods}.

Our experimental results prove that by training only one network, it is possible to learn new classes without catastrophically forgetting the previous classes. To the best of our knowledge, this is the first work, which proposes a solution for the incremental semantic segmentation of remote sensing data. 

\section{Methodology}

\subsection{Network Architecture}
Our network (see Fig.~\ref{fig:network_structure}) is a variant of U-net, which consists of an encoder that is architecturally the same as the first 13 convolutional layers of VGG16 \cite{simonyan2014very}, a corresponding decoder, mapping low resolution encoder feature maps to original input image size of outputs, and two center convolutional layers. We prefer to use VGG16 as the encoder, because it provides a good compromise between complexity and performance, as it is not as deep as e.g., VGG19 but still it is one of the best performers on famous benchmark challenges (e.g., ImageNet \cite{russakovsky2015imagenet}).

\begin{figure*}
	\centering
	\includegraphics[width=1\linewidth]{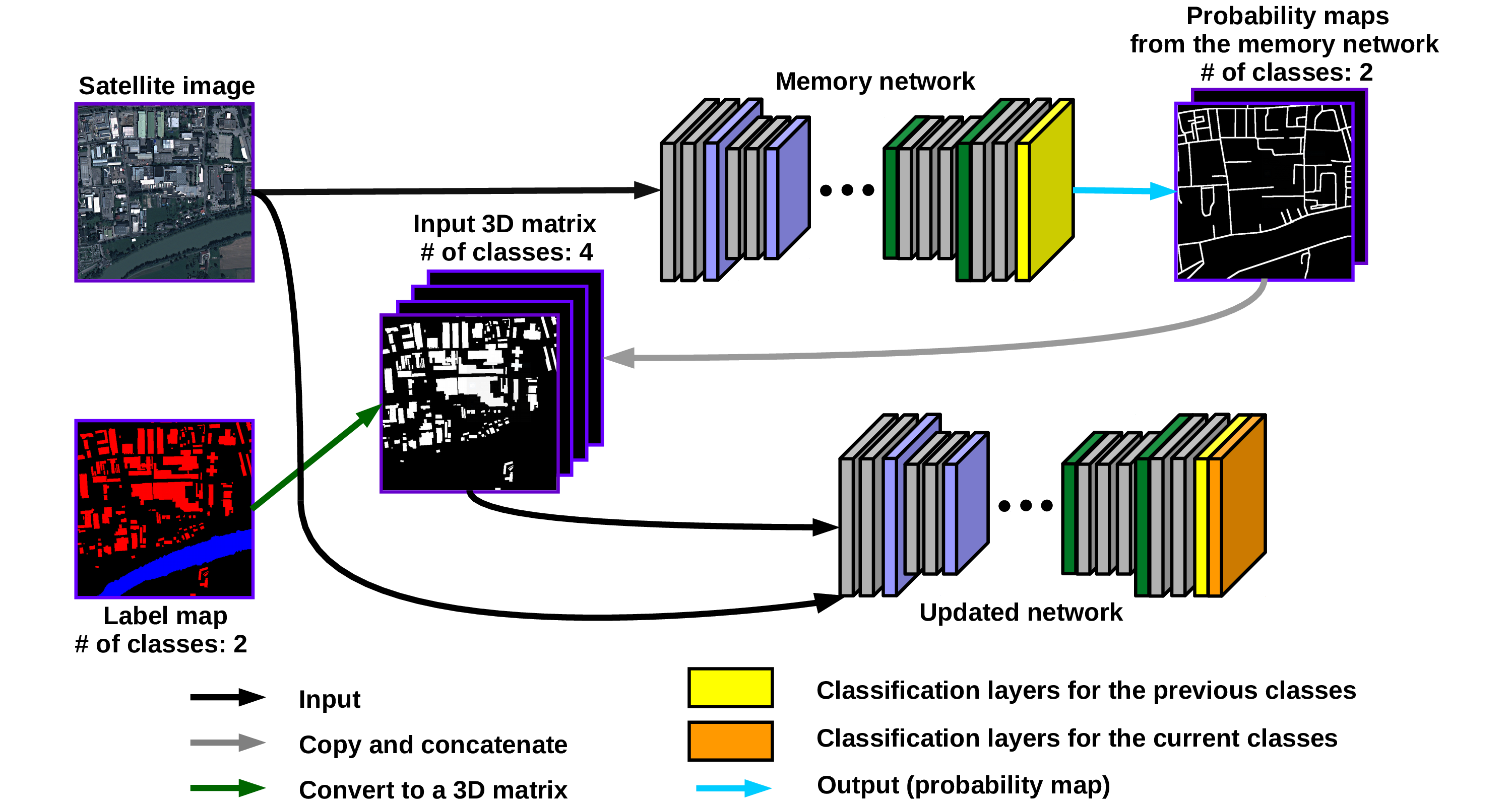}
	\caption{Adapting the network to the new training data. Although annotations for only 2 classes are provided, the updated network is still able to learn current classes as well as the previously learned 2 classes with the help of the memory network.}
	\label{fig:adaptation}
\end{figure*}

Output of each pooling layer in the encoder is concatenated with output of the symmetric deconvolutinal layer in the decoder through skip connections to combine higher level features with the lower ones. Kernel size and stride in all the convolutional layers are 3 and 1 respectively. Padding parameter in the convolutional layers is set to 1 so as to keep height and width of output the same as output of the previous layer. The max-pooling layers, having $2 \times 2$ window with stride 2 are used to halve width and height of the previous layer. In order to upsample output of the previous layer by factor of 2, both kernel size and stride parameters are set to 2 in deconvolutional layers. Except the last convolutional layer, all the convolution and deconvolution operations are followed by a ReLU. Since we conduct experiments on the Luxcarta dataset, containing images collected from different cities, instead of using batch normalization, which uses memory inefficiently, we prefer to add more patches in a batch.

Multi-task learning is the learning strategy, which solves multiple problems at the same time by learning all the tasks jointly. In deep neural networks, bottom layers enable to share information for all the tasks, whereas the last layers are dedicated to provide a solution for each task \cite{caruana1997multitask, ruder2017overview}. In incremental semantic segmentation problem, since the label maps of a remote sensing image for a class or several classes come sequentially, we consider the segmentation tasks as a multi-task learning problem, where performing a binary classification for each class corresponds to a different task. Output of our network is a 3D matrix that is a stack of binary predicted maps for all the classes. In the test stage, to generate a binary segmentation for each class, we first convert outputs of the final convolutional layers to probability maps using sigmoid; then, we threshold the probabilities at $0.5$.

\subsection{Adapting the Network to the New Training Data}
Fig.~\ref{fig:adaptation} depicts how the network is adapted to the new data.

Let us assume that the current training data are indicated by $D_{curr}$. We denote sets of the previously learned classes and the classes in $D_{curr}$ by $\mathcal{L}_{prev}$ and $\mathcal{L}_{curr}$, where $\mathcal{L}_{prev} \cap \mathcal{L}_{curr} = \emptyset$. 

The main goals we try to achieve during adaptation are to update the formerly trained network so that segmentation capabilities for $\mathcal{L}_{curr}$ are added, and to fine-tune the network on $D_{curr}$ for $\mathcal{L}_{prev}$, although annotations for $\mathcal{L}_{prev}$ are not available in $D_{curr}$. Output of the updated network is the matrix, consisting of binary segmentations for $\mathcal{L}_{updated} = \mathcal{L}_{prev} \cup \mathcal{L}_{curr}$.

\begin{figure*}
	\centering
	\includegraphics[width=0.8\linewidth]{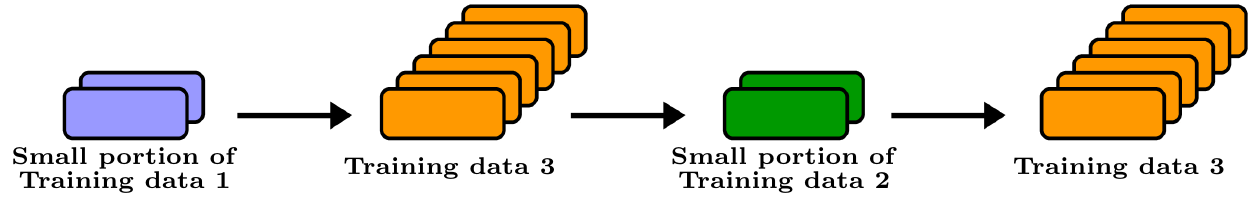}
	\caption{An example optimization sequence. The new classes are added on Training data~3 to the network, which was already trained on Training data~1 and Training data~2. The optimization sequence is as follows: $L_{rem}$ on Training data~1, $L_{adapt}$ on Training data~3, $L_{rem}$ on Training data~2, and $L_{adapt}$ on Training data~3 again.}
	\label{fig:optimization_sequence}
\end{figure*}

We use the knowledge distillation from the previously trained memory network as a proxy in absence of the ground-truth for $\mathcal{L}_{prev}$ in $D_{curr}$. We create an updated network, having exactly the same structure except the last classification layer, which has $|\mathcal{L}_{updated}|$ filters instead of $|\mathcal{L}_{prev}|$. During creation of the updated network, additional $|\mathcal{L}_{curr}|$ filters in the last classification layer are initialized using Xavier initialization \cite{glorot2010understanding}, and the rest of the parameters are loaded from the memory network. When $D_{curr}$ arrive, the incoming label map is first converted to a 3D matrix, consisting of binary ground-truth for $\mathcal{L}_{curr}$. The probability maps generated by the memory network are concatenated with this 3D matrix to provide information about $\mathcal{L}_{prev}$ to the updated network. The final 3D matrix as well as the input image in $D_{curr}$ are fed to the network as the new training data. While concatenating output of the memory network with the new ground-truth, we prefer to use soft probability maps generated by the memory network rather than hard classification maps in order to reduce the propagated error rate, caused by imprecision in output of the memory network, at each time step of incremental learning.

Let us denote the binary target label vectors for $n$ training samples $i = 1 \ldots n$ in a batch from $D_{curr}$ by $\mathbf{y}_{curr}^{(i)}$ and the predicted probabilities for $\mathcal{L}_{prev}$ from the memory network by $\mathbf{\hat{{y}}}^{(i)}_{mem}$. We denote by $\mathbf{\hat{{y}}}^{(i)}_{up\_curr}$ and $\mathbf{\hat{{y}}}^{(i)}_{up\_prev}$, the predicted probabilities for $\mathcal{L}_{curr}$ and $\mathcal{L}_{prev}$ from the updated network. The classification loss $L_{class}$ quantifies mismatch between $\mathbf{y}_{curr}^{(i)}$ and $\mathbf{\hat{{y}}}^{(i)}_{up\_curr}$. In order to compute $L_{class}$, since we deal with generation of a binary segmentation for each class as a separate task, we use sigmoid cross entropy loss defined as:
\begin{equation}\label{eq:class_loss}
\begin{split}
L_{class} = - \frac{1}{n{|\mathcal{L}_{curr}|}} \sum_{i = 1}^{n} \sum_{k = 1}^{|\mathcal{L}_{curr}|} \Big[ y_{curr(k)}^{(i)} \text{log} \left( \hat{y}^{(i)}_{up\_curr(k)} \right) + \\
\left( 1 - y_{curr(k)}^{(i)} \right) \text{log} \left( 1 - \hat{y}^{(i)}_{up\_curr(k)} \right) \Big].
\end{split}
\end{equation}
In order for the updated network to learn $\mathcal{L}_{prev}$ on $D_{curr}$, we try to keep discrepancy between $\mathbf{\hat{{y}}}^{(i)}_{up\_prev}$  and $\mathbf{\hat{{y}}}^{(i)}_{mem}$ as small as possible. The distillation loss $L_{distil}$, which measures this disparity is defined as:
\begin{equation}\label{eq:distil_loss}
\begin{split}
L_{distil} = - \frac{1}{n{|\mathcal{L}_{prev}|}} \sum_{i = 1}^{n} \sum_{k = 1}^{|\mathcal{L}_{prev}|} \Big[ \hat{{y}}^{(i)}_{mem(k)} \text{log} \left( \hat{y}^{(i)}_{up\_prev(k)} \right) + \\
\left( 1 - \hat{{y}}^{(i)}_{mem(k)} \right) \text{log} \left( 1 - \hat{y}^{(i)}_{up\_prev(k)} \right) \Big].
\end{split}
\end{equation}
The overall adaptation loss $L_{adapt}$ that is optimized during adaptation is computed by adding these two terms:
\begin{equation}\label{eq:adapt_loss}
L_{adapt} = L_{class} + L_{distil}.
\end{equation}
\subsection{Remembering From the Previous Training Data}\label{sec:remembering_from_the_old_data}
We denote the previous training data by $D_{prev} = D_{prev}^{(1)} \cup D_{prev}^{(2)} \cup \ldots \cup D_{prev}^{(m)}$, where $D_{prev}^{(1)}$ corresponds to the first data, $D_{prev}^{(2)}$ is the second data, and so forth. If the training data are captured sequentially from different geographic locations, in order for the network not to overfit on $D_{curr}$ for $\mathcal{L}_{prev}$, we remind the previously learned information by systematically showing patches from the stored, small portion of $D_{prev}$. Since in most of the cases classes in the training data are highly imbalanced, when determining which training patches to store in $D_{prev}^{(j)}$, random selection may cause storing no samples for less frequent classes. For this reason, we take the class imbalance problem into account. We first compute weight $w_{c}$ of each class $c \in \mathcal{L}_{prev}^{(j)}$ in $D_{prev}^{(j)}$ as:
\begin{equation}
w_{c} = \frac{\text{median} ({f_{c} | c \in \mathcal{L}_{prev}^{(j)}})}{f_{c}},
\end{equation}\label{eq:mfb}
where $f_{c}$ denotes frequency of the pixels that are labeled as class $c$. We then assign an importance value $I^{(l)}$ to the $l^{th}$ training patch in $D_{prev}^{(j)}$ as:
\begin{equation}
I^{(l)} = \sum_{c \in \mathcal{L}_{prev}^{(j)}}w_{c}f_{c}^{(l)},
\end{equation}
where $f_{c}^{(l)}$ denotes the number of pixels, belonging to $c$ in the patch. We store certain number of patches that have the highest $I$ value, which we denote by $D_{prev\_imp}^{(j)}$. In order to diversify the patches that are fed to the network, we randomly select a small fraction of the remaining patches. We denote the randomly chosen patches by $D_{prev\_random}^{(j)}$. The data to be stored from $D_{prev}^{(j)}$ for remembering are $D_{prev\_rem}^{(j)}~=~D_{prev\_imp}^{(j)}~\cup~D_{prev\_random}^{(j)}$. The number of patches that is selected randomly and using the importance value needs to be determined by the end user.

Let us denote the target vector for the $i^{th}$ sample among $n$ samples in a batch from $D_{prev\_rem}^{(j)}$ by $\mathbf{y}^{(j)(i)}_{prev}$. We denote by $\hat{y}^{(j)(i)}_{up\_prev}$ the predicted vector from the updated network for the same sample. The remembering loss $L_{rem}$ is calculated as:
\begin{equation}\label{eq:remembering_loss}
\begin{split}
L_{rem} = - \frac{1}{n{|\mathcal{L}_{prev}^{(j)}|}} \sum_{i = 1}^{n} \sum_{k = 1}^{|\mathcal{L}_{prev}^{(j)}|} \Big[ {y}^{(j)(i)}_{prev(k)} \text{log} \left( \hat{y}^{(j)(i)}_{up\_prev(k)} \right) + \\
\left( 1 - {y}^{(j)(i)}_{prev(k)} \right) \text{log} \left( 1 - \hat{y}^{(j)(i)}_{up\_prev(k)} \right) \Big].
\end{split}
\end{equation}
During remembering from $D_{prev}^{(j)}$, we freeze the classification layers that are responsible for $c \not\in \mathcal{L}_{prev}^{(j)}$ and optimize the rest of the network. The user needs to determine how often and on which data $L_{rem}$ is optimized. An example optimization sequence is depicted in Fig.~\ref{fig:optimization_sequence}.
\section{Experiments}\label{sec:experiments}
\subsection{Methods Used for Comparison}\label{sec:compared_methods}
Table~\ref{table:advantages_disadvantages} compares our methodology with the following approaches:
\begin{table*}
\caption{Advantages and disadvantages of our approach with respect to the compared methods.}
\label{table:advantages_disadvantages}
\centering
\begin{tabular}{c|c|c|c|c|c|c} 
\hline
\multirow{3}{*}{\textbf{Method}} 
 & \textbf{Training}                      & \multirow{2}{*}{\textbf{Test}} & \textbf{Performance}  & \textbf{Performance}              & \textbf{Convergence}              & \multirow{2}{*}{\textbf{Number of}}   \\
 & \textbf{Time}                          & \multirow{2}{*}{\textbf{Time}} & \textbf{for the new}  & \textbf{for the old}              & \textbf{time for the new}         & \multirow{2}{*}{\textbf{Classifiers}} \\
 & \textbf{(1 iteration)}                 &                                & \textbf{classes}      & \textbf{classes}                  & \textbf{classes}                  & \\
\hline              
\multirow{2}{*}{\textit{static learning}} & \multirow{2}{*}{fast}          & \multirow{2}{*}{fast} & \textcolor{red}{continual learning} & \textcolor{red}{continual learning} & \textcolor{red}{continual learning} & \multirow{2}{*}{1}\\
                                          &                                &                       & \textcolor{red}{is not supported} & \textcolor{red}{is not supported} & \textcolor{red}{is not supported}    & \\
\hline
\textit{multiple learning}                & fast                           & \textcolor{red}{very slow} & good                         & good                              & \textcolor{amber}{medium}         & \textcolor{red}{N} \\
\hline  
\textit{fixed representation}             & \textcolor{green}{very fast}   & fast                       & \textcolor{red}{very bad}    & good                              & \textcolor{red}{can not learn}    & 1 \\
\hline 
\textit{fine-tuning}                      & fast                           & fast                       & good                         & \textcolor{red}{very bad}         & very fast                         & 1 \\
\hline
\textit{incremental learning}             & \textcolor{amber}{medium}      & \textcolor{green}{fast}    & \textcolor{green}{good}      & \textcolor{green}{good}           & \textcolor{green}{very fast}      & \textcolor{green}{1} \\
                        
\hline
\end{tabular}
\end{table*}

\textit{Static learning: } This is the traditional learning approach, where we assume that all the training images and annotations for the same classes are available at the time of training. In real-world segmentation problems, this condition is extremely hard to meet. This method does not support learning new classes continually. 

\textit{Multiple learning: } In this learning strategy, we train an additional classifier whenever the new training data are obtained. The number of classifiers that needs to be stored increases linearly. In addition, because the test images have to be segmented using all the trained classifiers to generate a map for each class, the test stage might be extremely long. Therefore, this approach is extremely expensive in terms of storage and segmentation efficiency.

\textit{Fixed representation: } To learn new classes, we remove the classification layers, which were optimized for the previous classes and plug in new classification layers dedicated for the new classes. The newly added classification layers are initialized with Xavier method \cite{glorot2010understanding}. When new training data arrive, we optimize only the newly added classification layers and freeze the rest of the network. Hence, training is very fast. During testing, we append the formerly trained classification layers back to the network to generate label maps for all the classes. The major issue is that although performance for the initial classes is preserved, the network struggles in learning new classes, because the previously extracted features are not optimized to represent the new classes.

\begin{figure}
\centering
\includegraphics[width=0.85\linewidth]{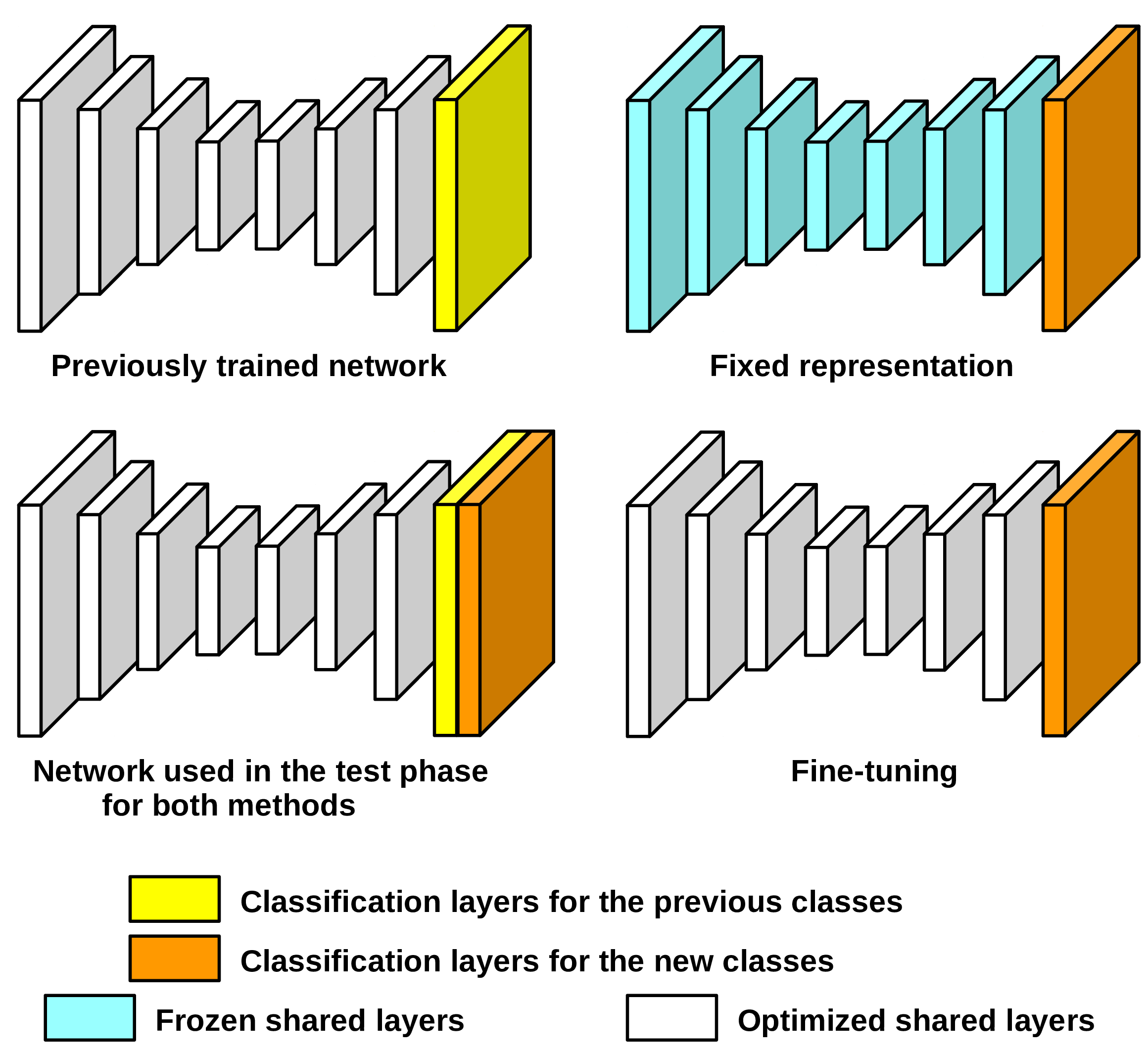}
\caption{Example network structures for \textit{fixed representation} and \textit{fine-tuning}. During the test stage, classification layers for the previous classes are appended to the network to generate label maps for all the classes.}
\label{fig:ft_fe_illustration}
\end{figure}

\textit{Fine-tuning: } We use a similar strategy that we follow in \textit{fixed representation}. The only difference is that while training the network, instead of only the classification layers, we optimize the whole network using only the new training data. In this methodology, although the network performs a remarkable performance for the new classes, it suffers from catastrophic forgetting. Example network structures for \textit{fixed representation} and \textit{fine-tuning}, for both training and test phases, are illustrated in Fig.~\ref{fig:ft_fe_illustration}.

For \textit{incremental learning}, it is required for the memory network to generate probability maps from the training patches to optimize $L_{distil}$. Therefore, training time for our approach is slightly longer than the others. This can be considered as the only disadvantage of the proposed methodology.

\begin{figure}
\centering
\subfloat[Albi]{\includegraphics[width=0.33\linewidth]{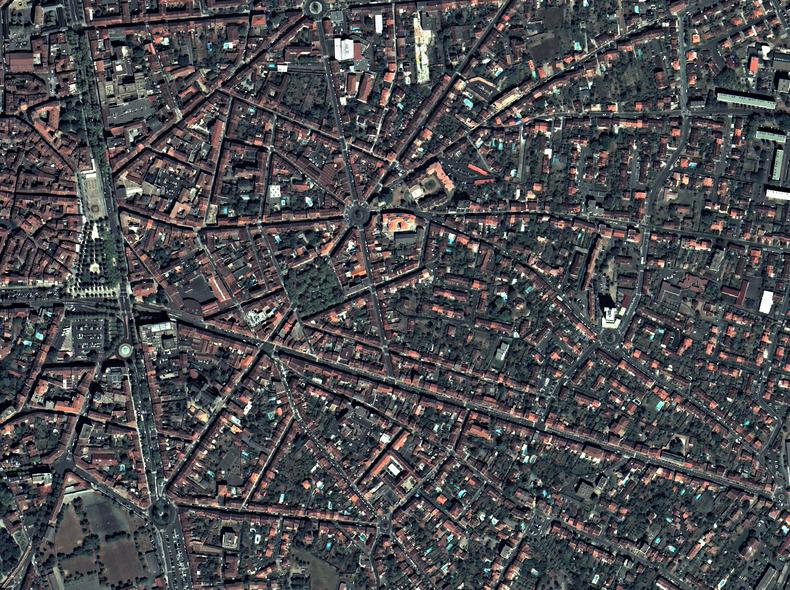}}
\hfill
\subfloat[Enns]{\includegraphics[width=0.33\linewidth]{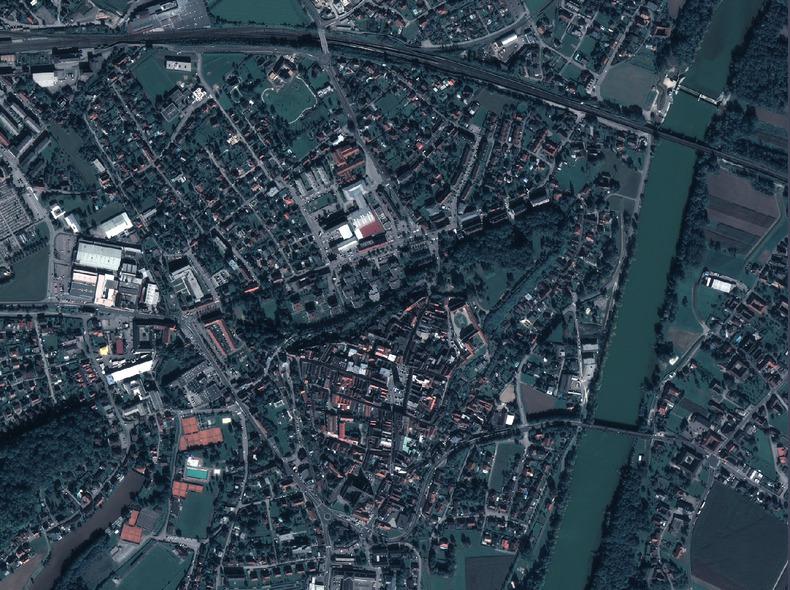}}
\hfill
\subfloat[Lyon]{\includegraphics[width=0.33\linewidth]{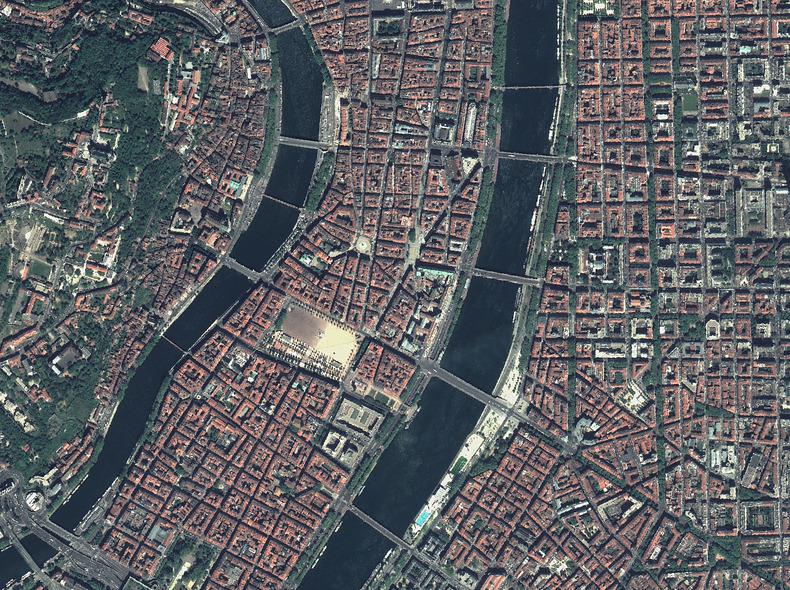}}
\caption{Example close-ups from the Luxcarta dataset.}
\label{fig:lux_data_ex_images}
\end{figure}

\subsection{Datasets and Evaluation Metrics}
The first data we use are the Luxcarta dataset, containing 8 bit satellite images collected from 22 different cities in Europe. 11 of these cities are located in France and the other 11 are in Austria, covering the total area of approximately 1367 km$^{2}$. The images were collected from the following cities: \textit{Amstetten}, \textit{Enns}, \textit{Leibnitz}, \textit{Salzburg}, \textit{Villach}, \textit{Bad Ischl}, \textit{Innsbruck}, \textit{Klagenfurt}, \textit{Osttirol}, \textit{Sankt P{\"o}lten}, \textit{Voitsberg} in \textit{Austria}, and \textit{Albi}, \textit{Angers}, \textit{Bayonne-Biarritz}, \textit{B{\'e}ziers}, \textit{Bourges}, \textit{Douai}, \textit{Draguignan}, \textit{Lille}, \textit{Lyon}, \textit{N{\^{i}}mes}, \textit{Roanne} in \textit{France}. Spectral bands of the images are composed of Red (R), Green (G), and Blue (B) channels. The spatial resolution is 1 m. Since the images were captured over different geographic locations, they have different color distributions and visual features. Example close-ups from the dataset are shown in Fig.~\ref{fig:lux_data_ex_images}. Annotations for \textit{building}, \textit{road}, \textit{high vegetation}, \textit{water}, and \textit{railway} classes are provided as shape files. \textit{Road} class has 4 sub-classes: \textit{main road}, \textit{secondary road}, \textit{street}, and \textit{highway}. The ground-truth for \textit{road} and \textit{railway} classes are polylines, and the label maps for the other classes are polygones. To have full annotations, we raster all the shape files using \textit{GDAL}\footnote{https://www.gdal.org}. We enlarge \textit{highways} and \textit{secondary roads} by applying morphological dilation with 8 $\times$ 8 and 4 $\times$ 4 square shaped structuring element (SE), respectively. \textit{Railways} and other type of \textit{roads} are enlarged using the same method with 6 $\times$ 6 SE. Size of the SEs is determined by visual inspection.

The other two datasets, on which we conduct our experiments are the Vaihingen and the Potsdam benchmarks provided by the ISPRS \cite{isprs_data}. Both datasets contain 8 bit aerial images.  The Vaihingen dataset consists of 33 image tiles (of average size 2494 $\times$ 2064), where 16 of them are provided as training and the rest as test. The images comprise 3 spectral bands: Near Infrared (NIR), R, and G. The spatial resolution is 9 cm. The Potsdam dataset includes 38 tiles (of size 6000 $\times$ 6000), out of which 24 are dedicated for training and the remaining for test. The images contain 5 channels: NIR, R, G, B, and the normalized DSM (nDSM) data. The resolution of the images in this benchmark is 5 cm. Both datasets contain full annotations for 6 classes: \textit{impervious surfaces}, \textit{building}, \textit{low vegetation}, \textit{high vegetation}, \textit{car}, and \textit{clutter}. However, since only 0.78$\%$ of the pixels in the Vaihingen dataset is labeled as \textit{clutter}, we ignore this class in the experiments on this benchmark. As of 2018 summer, the competition for these benchmarks is over, and all the reference data are publicly available. Hence, we use all the training tiles for training, and test tiles for validation. To account for the labeling mistakes while the datasets are annotated, the eroded ground-truth is also provided. We use this ground-truth to assess the performance on the benchmarks.

To quantitatively assess the performance for each class, we compare the binary predicted map and the binary ground-truth using two evaluation metrics: intersection over union (IoU)~\cite{csurka2013good} and F1-score~\cite{audebert2016semantic}.

\subsection{Experiments on the Luxcarta Dataset}\label{sec:experiments_on_luxcarta}
In this experimental setup, we suppose that the training data are obtained sequentially in time, and every snapshot of the streaming training data contains the satellite images from different cities and label maps for separate classes. As the training data, we use 18 cities, out of which 9 are located in \textit{Austria} and the other 9 are in \textit{France}. We use 2 cities from each country for validation. We split the training cities into three sets as reported in Table~\ref{table:luxcarta_training_validation_cities} by paying attention that the cities in each set are the ones, which contain a reasonable amount of samples for the given annotations, and whose color distributions are as diverse as possible. We assume that the training cities are streamed in this order: Train1, Train2, Train3. For \textit{multiple learning}, \textit{fixed representation}, and \textit{fine-tuning} we assume that the previous data are not accessible. For \textit{incremental learning}, we store only $30\%$ of the training patches in the previous data, out of which $15\%$ are selected using the importance value and $15\%$ are chosen randomly, as explained in section \ref{sec:remembering_from_the_old_data}. We also test our approach without accessing the previous data (i.e., without optimizing $L_{rem}$), which we refer as \textit{incremental learning w/o $L_{rem}$}. Since \textit{static learning} does not support adding new classes continually, for this approach, we assume that the training images from 18 different cities and label maps of all 5 classes for each image are available beforehand to train a single network. For this reason, we expect it to be an obvious upper bound of the other methods. 

\begin{table}
\centering
\caption{Training and validation cities of the Luxcarta dataset.}
\label{table:luxcarta_training_validation_cities}
\begin{tabular}{||c|c||c|c||c||}
\hline
\multicolumn{2}{||c||}{\textbf{Data Type and}} & \multicolumn{2}{|c||}{\textbf{Data Type and}} & \multirow{6}{*}{\textbf{City (Country)}}\\
\multicolumn{2}{||c||}{\textbf{Classes for}} & \multicolumn{2}{|c||}{\textbf{Classes for}} & \\
\cline{1-4}
\multicolumn{2}{||c||}{\textit{mult. learning}}    & \multicolumn{2}{|c||}{\multirow{4}{*}{\textit{static learning}}} & \\
\multicolumn{2}{||c||}{\textit{fixed rep.}}        & \multicolumn{2}{|c||}{}                                & \\
\multicolumn{2}{||c||}{\textit{fine-tuning}}       & \multicolumn{2}{|c||}{}                                   & \\
\multicolumn{2}{||c||}{\textit{inc. learning}}     & \multicolumn{2}{|c||}{}                                   & \\
\hline
\multirow{7}{*}{\rotatebox[origin=c]{90}{Train1}} & \multirow{4}{*}{building} & 
\multirow{18}{*}{\rotatebox[origin=c]{90}{Train}}  &                         & \textit{Bad Ischl} (\textit{Austria})        \\
                                 &                            & &                         & \textit{Osttirol} (\textit{Austria})         \\
                                 &                            & &                         & \textit{Voitsberg} (\textit{Austria})        \\
                                 &                            & &                         & \textit{Bayonne-Biarritz} (\textit{France})  \\
                                 & \multirow{3}{*}{high veg.} & &                         & \textit{Bourges} (\textit{France})           \\
                                 &                            & &    building             & \textit{Draguignan} (\textit{France})        \\
                                 &                            & &                         & \textit{N{\^{i}}mes} (\textit{France})       \\
\cline{1-2}
\cline{5-5}
\multirow{6}{*}{\rotatebox[origin=c]{90}{Train2}} &  \multirow{3}{*}{road} &    &    high veg.            & \textit{Enns} (\textit{Austria})             \\
                                 &                            & &                         & \textit{Innsbruck} (\textit{Austria})        \\
                                 &                            & &    road                 & \textit{Klagenfurt} (\textit{Austria})       \\
                                 &  \multirow{3}{*}{railway}  & &                         & \textit{Sankt P{\"o}lten} (\textit{Austria}) \\
                                 &                            & &    railway              & \textit{B{\'e}ziers} (\textit{France})       \\
                                 &                            & &                         & \textit{Lyon} (\textit{France})              \\
\cline{1-2}
\cline{5-5}
\multirow{5}{*}{\rotatebox[origin=c]{90}{Train3}} & \multirow{5}{*}{water} &     &    water                & \textit{Albi} (\textit{France})              \\
                                 &                            & &                         & \textit{Villach} (\textit{Austria})          \\
                                 &                            & &                         & \textit{Salzburg} (\textit{Austria})         \\
                                 &                            & &                         & \textit{Angers} (\textit{France})            \\
                                 &                            & &                         & \textit{Douai} (\textit{France})            \\
\hline
\multirow{5}{*}{\rotatebox[origin=c]{90}{Validation}}  
& building  & \multirow{5}{*}{\rotatebox[origin=c]{90}{Validation}} & building  & \multirow{2}{*}{ \textit{Amstetten} (\textit{Austria})} \\
& high veg. &                                                       & high veg. & \multirow{2}{*}{\textit{Leibnitz} (\textit{Austria})}   \\
& road      &                                                       & road      & \multirow{2}{*}{ \textit{Lille} (\textit{France})}   \\
& railway   &                                                       & railway   & \multirow{2}{*}{ \textit{Roanne} (\textit{France})}     \\
& water     &                                                       & water     &                               \\
\hline
\end{tabular}
\end{table}

During the pre-processing step, we split all the training images into 384~$\times$~384 patches with an overlap of 32~$\times$~32 pixels between the neighboring patches. The validation images are divided into 2240~$\times$~2240 patches with 64~$\times$~64 pixels of overlap. After all the validation patches are classified, they are combined back to get the original size classification maps. Because the satellite images arrive sequentially (except for \textit{static learning}), it is not possible to compute mean values for the image channels. Hence, for the normalization, we subtract 127 from all the pixels, as the images are 8 bit.

\begin{table*}
\centering
\caption{F1 scores on the Luxcarta dataset.}
\label{table:luxcarta_f1}
\begin{tabular}{|c|cccccc||c||}
\hline
\textbf{Method} & \textbf{Epoch} & \textbf{Building} & \textbf{High veg.} & \textbf{Road} & \textbf{Railway} & \textbf{Water} & \textbf{Overall} \\
\hline
\textit{static learning} 
 & \textbf{500}  & \multicolumn{1}{||c}{80.74 (Ref.)} & 71.26 (Ref.)                       & 66.21 (Ref.) & 61.72 (Ref.)                      & 82.74 (Ref.)        & 72.54 (Ref.)       \\
\hline
\hline
\textit{multiple learning}
 & \textbf{500}  & \multicolumn{1}{||c}{71.25} & \multicolumn{1}{c||}{68.88} & 59.28 & \multicolumn{1}{c||}{\textbf{55.65}} & 79.83        &  66.98 (-5.56)  \\
\hline 
\multirow{2}{*}{\textit{fixed representation}}
 & 1000          & \multicolumn{1}{||c}{71.25} & \multicolumn{1}{c||}{68.88} & 2.71  & \multicolumn{1}{c||}{0.00}  & \textemdash  &   \\
 & \textbf{1500} & \multicolumn{1}{||c}{71.25} & \multicolumn{1}{c||}{68.88} & 2.71  & \multicolumn{1}{c||}{0.00}  & 0.11         & 28.59 (-43.95)\\
\hline 
\multirow{2}{*}{\textit{fine-tuning}}
 & 1000          & \multicolumn{1}{||c}{28.91} & \multicolumn{1}{c||}{0.17}  & 59.30 & \multicolumn{1}{c||}{60.06} & \textemdash  &   \\
 & \textbf{1500} & \multicolumn{1}{||c}{27.90} & \multicolumn{1}{c||}{7.71}  & 0.14  & \multicolumn{1}{c||}{0.01}  & \textbf{90.20}        &  25.19 (-47.35) \\
\hline  
\textit{incremental learning}
 & 1000          & \multicolumn{1}{||c}{74.19} & \multicolumn{1}{c||}{66.32} & 56.57 & \multicolumn{1}{c||}{50.87} & \textemdash  &   \\
\textit{w/o $L_{rem}$}
 & \textbf{1500} & \multicolumn{1}{||c}{74.91} & \multicolumn{1}{c||}{66.87} & 58.14 & \multicolumn{1}{c||}{51.70} & 82.32        &  66.79 (-5.75)\\
\hline
\multirow{2}{*}{\textit{incremental learning}}
 & 1000          & \multicolumn{1}{||c}{75.98} & \multicolumn{1}{c||}{72.38}  & 57.29 & \multicolumn{1}{c||}{50.18} & \textemdash  &   \\
 & \textbf{1500} & \multicolumn{1}{||c}{\textbf{76.78}} & \multicolumn{1}{c||}{\textbf{72.06}}  & \textbf{59.58} & \multicolumn{1}{c||}{53.07} & 78.94       & \textbf{68.09 (-4.45)}\\
 
\hline
\multicolumn{1}{c}{} &  & \multicolumn{2}{||c||}{\textbf{Training Set 1}}  & \multicolumn{2}{|c||}{\textbf{Training Set 2}} & \textbf{Training Set 3} & \multicolumn{1}{c}{}\\
\cline{3-7} 
\end{tabular}
\end{table*}

We train a single model for \textit{static learning} using the whole training data for 500 epochs, in which each epoch has 100 iterations. For \textit{multiple learning}, we train 3 separate models from scratch on Train1, Train2, and Train3 with the same hyper-parameters. For \textit{fixed representation}, \textit{fine-tuning}, and the proposed \textit{incremental learning} methodologies, every time when the new classes are added from the new data, we optimize the network for the same number of epochs and iterations as for \textit{static learning} and \textit{multiple learning}. In every 5 training iterations of the network for \textit{incremental learning} approach on Train2, we optimize $L_{rem}$ on Train1 for 1 iteration and $L_{adapt}$ for the next consecutive 4 iterations. During the training on Train3, since the network has already learned information from both Train1 and Train2, we prefer to remind the network the previously learned information more often. On Train3, the optimization sequence as follows: $L_{rem}$ on Train1 for 1 iteration, $L_{adapt}$ for 2 iterations, $L_{rem}$ on Train2 for 1 iteration, and $L_{adapt}$ for 2 iterations again.

\begin{figure}
\centering
\subfloat[$k = 1.4$]{\includegraphics[width=0.195\linewidth]{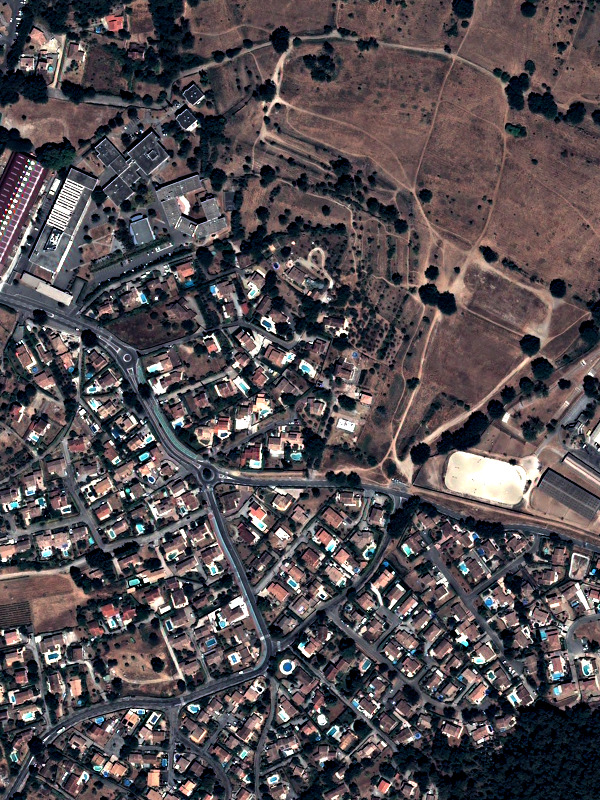}}
\hfill
\subfloat[$k = 0.6$]{\includegraphics[width=0.195\linewidth]{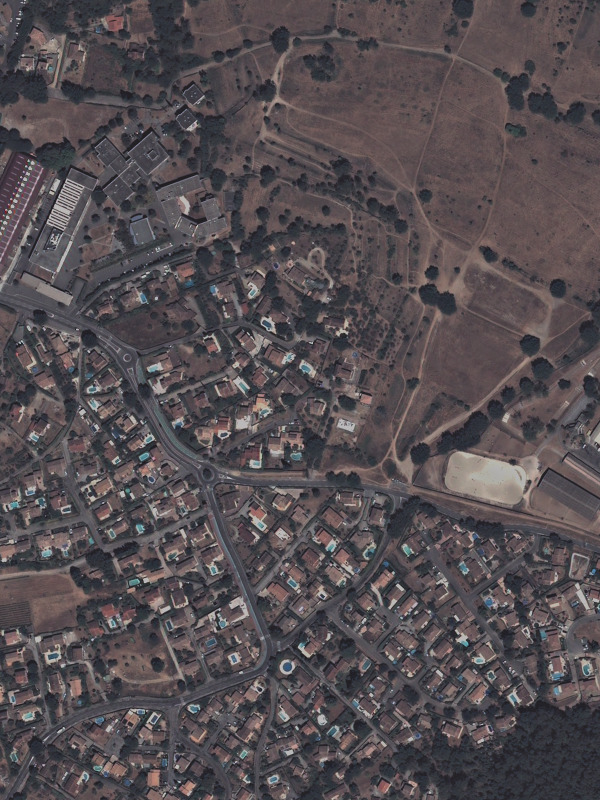}}
\hfill
\subfloat[Image]{\includegraphics[width=0.195\linewidth]{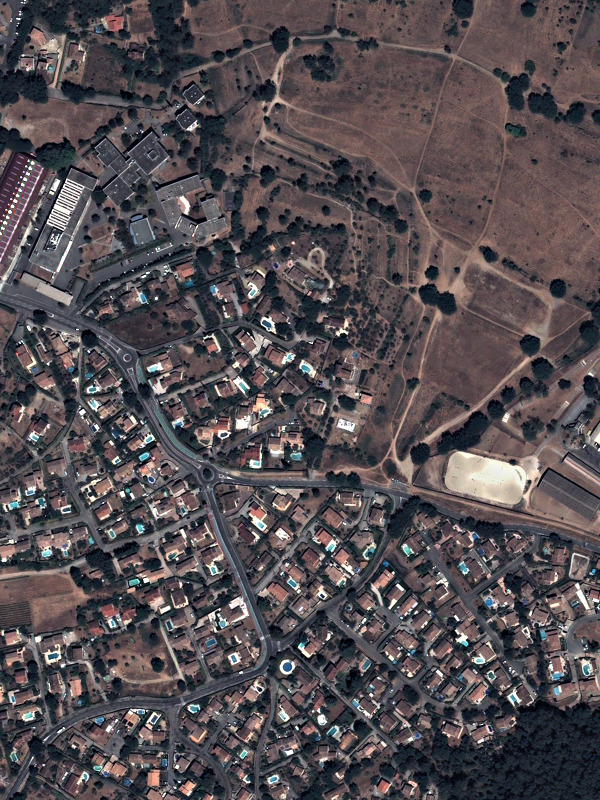}}
\hfill
\subfloat[$\gamma = 0.6$]{\includegraphics[width=0.195\linewidth]{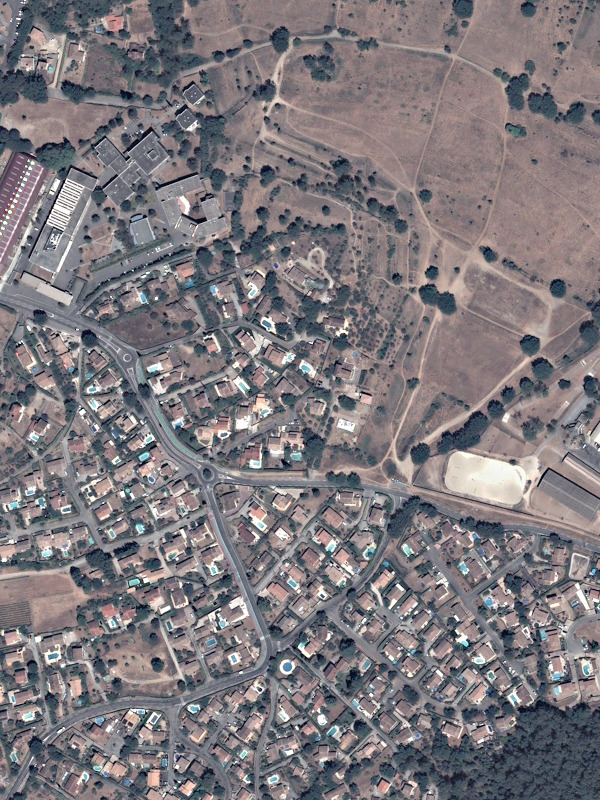}}
\hfill
\subfloat[$\gamma = 1.4$]{\includegraphics[width=0.195\linewidth]{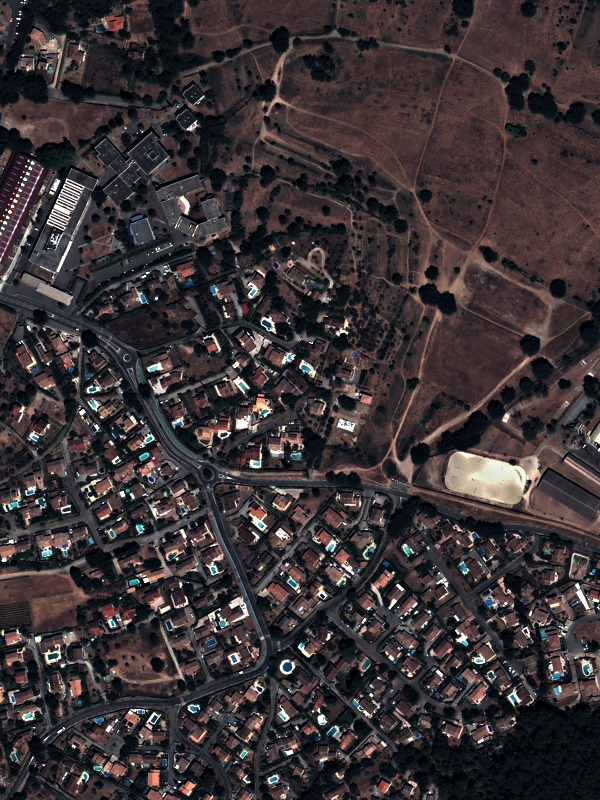}}
\caption{Illustration of the contrast change (a - b) and the gamma correction (d - e) for an example input image (c).}
\label{fig:data_augmentation}
\end{figure}

To update parameters of the network, we use \textit{Adam optimizer}, where the learning rate is $0.0001$, exponential decay rate for the first and the second moment estimates are $0.9$ and $0.999$, respectively. In every training iteration, a mini-batch of 12 patches is used for the optimization. When sampling a patch, we first select a random country (i.e., \textit{Austria} or \textit{France}). We then sample a random patch belonging to the city, which is also randomly chosen from the selected country. While training the network we apply online data augmentation to enrich the training data. The patches are augmented by random vertical/horizontal flips, 0/90/180/270 degrees of rotations, and distorting their radiometry by random contrast change and gamma correction. Contrast of each channel in the image is changed as:
\begin{equation}\label{eq:contrast_change}
x_{curr} = (x_{prev} - \mu) * k + \mu,
\end{equation}
where $x_{prev}$ and $\mu$ are the pixel value and mean of all the pixels before the change, $x_{curr}$ is the pixel value after the change, and $k$ is the distortion factor, for which we generate a random value between 0.75 and 1.5. Gamma correction is formulated as:
\begin{equation}\label{eq:gamma_correction}
x_{curr} = {x_{prev}}^{\gamma},
\end{equation}
where $\gamma$ is the correction factor, which is drawn uniformly between 0.75 and 1.25. In Eqs.~\eqref{eq:contrast_change} and \eqref{eq:gamma_correction}, we assume that the pixel values range between [0-1]. Fig.~\ref{fig:data_augmentation} illustrates the effect of gamma correction and the contrast change. 

\begin{figure*}
\centering
\subfloat[Static learning\label{fig:luxcarta_plots_joint}]{\includegraphics[width=0.49\linewidth]{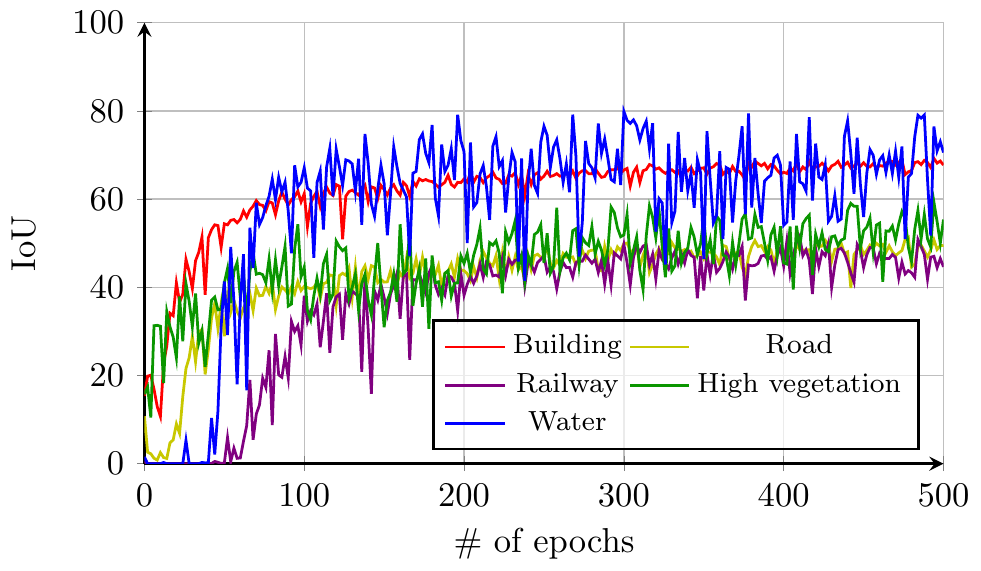}}
\hfill
\subfloat[Multiple learning\label{fig:luxcarta_plots_mult}]{\includegraphics[width=0.49\linewidth]{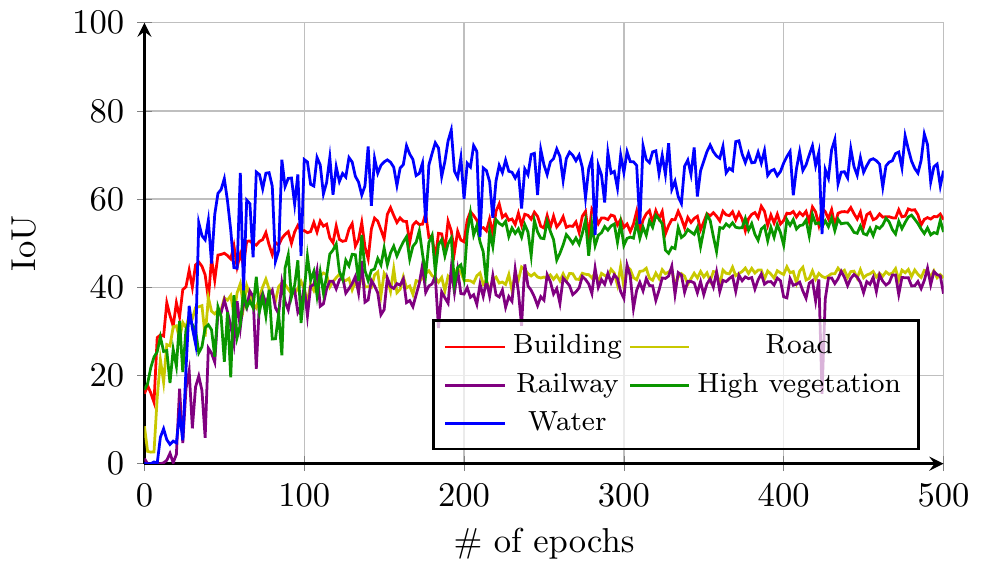}}
\hfill
\subfloat[Fixed representation\label{fig:luxcarta_plots_fe}]{\includegraphics[width=1\linewidth]{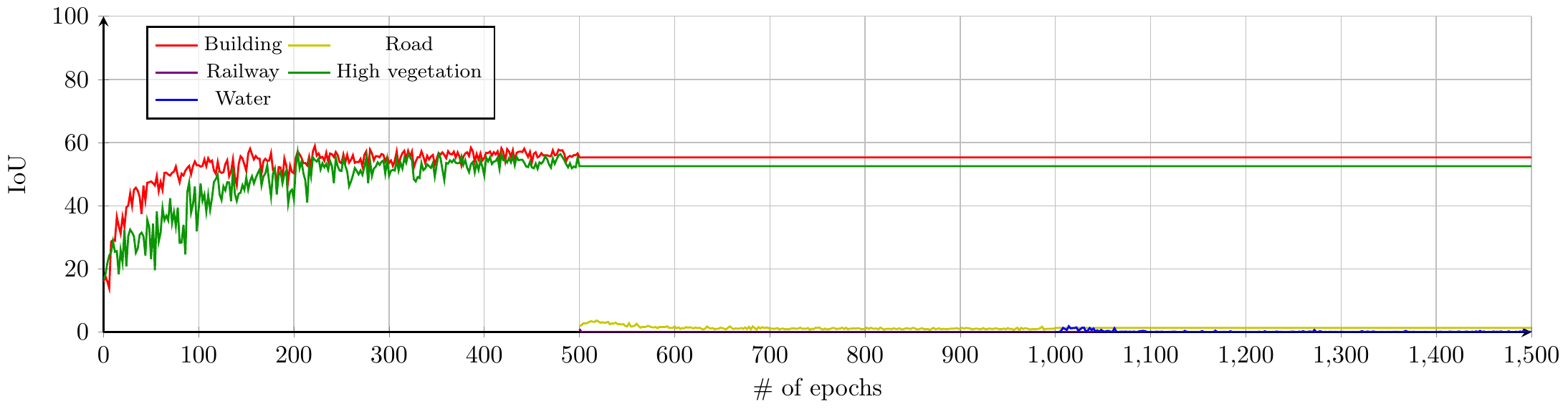}}
\hfill
\subfloat[Fine-tuning\label{fig:luxcarta_plots_ft}]{\includegraphics[width=1\linewidth]{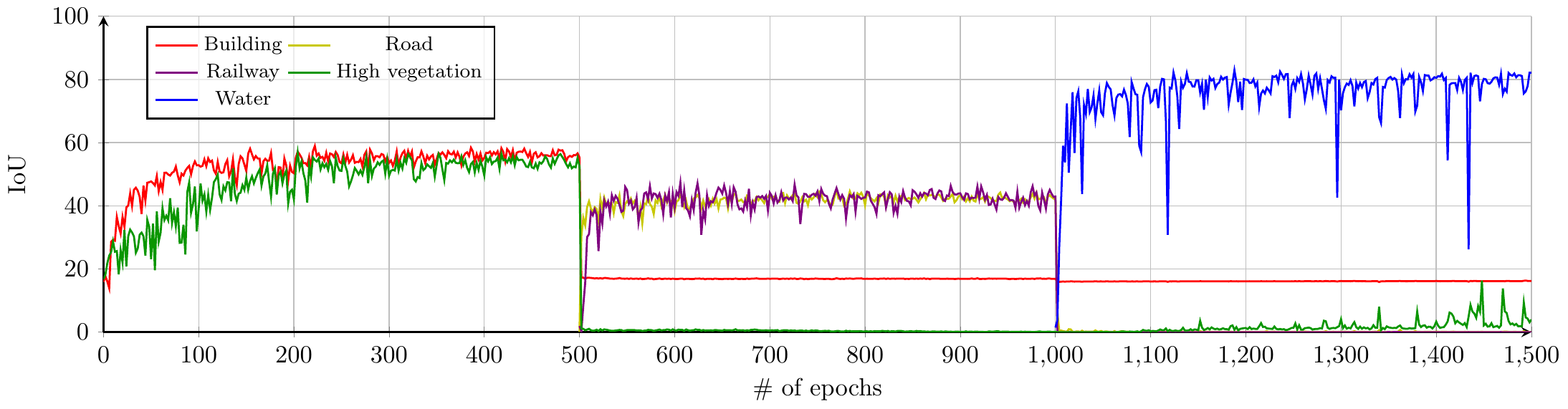}}
\hfill
\subfloat[Incremental learning\label{fig:luxcarta_plots_inc}]{\includegraphics[width=1\linewidth]{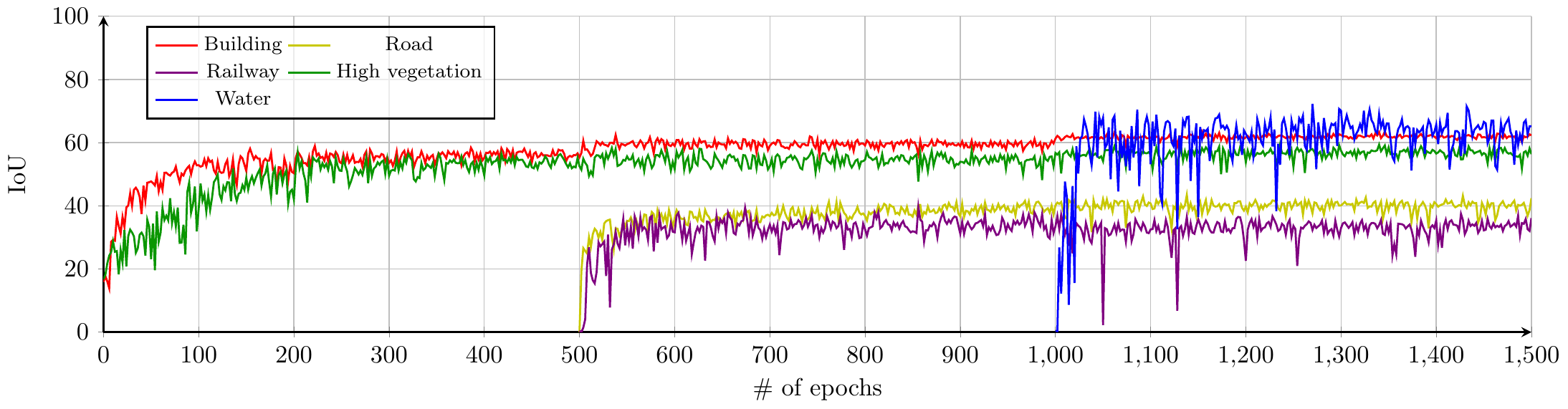}}
\hfill
\caption{Plots for the overall IoU values on the 4 validation cities of the Luxcarta dataset.}
\label{fig:luxcarta_plots}
\end{figure*}

\begin{figure*}
\centering
\begin{tabular}{p{0.1em}c@{\hspace{0.15em}}c@{\hspace{0.15em}}c@{\hspace{0.15em}}c@{\hspace{0.15em}}c@{\hspace{0.15em}}c}

& \multirow{2}{*}{Image} & \multirow{2}{*}{Ground-truth} & \multirow{2}{*}{Static learning} & \multirow{2}{*}{Multiple learning} & Inc. learning   & \multirow{2}{*}{Inc. learning} \\
&                        &                               &                                  &                                    & w/o $L_{rem}$ & \\
\rotatebox[origin=c]{90}{\textit{Amstetten}}&
\raisebox{-.5\height}{\frame{\includegraphics[width=0.16\linewidth]{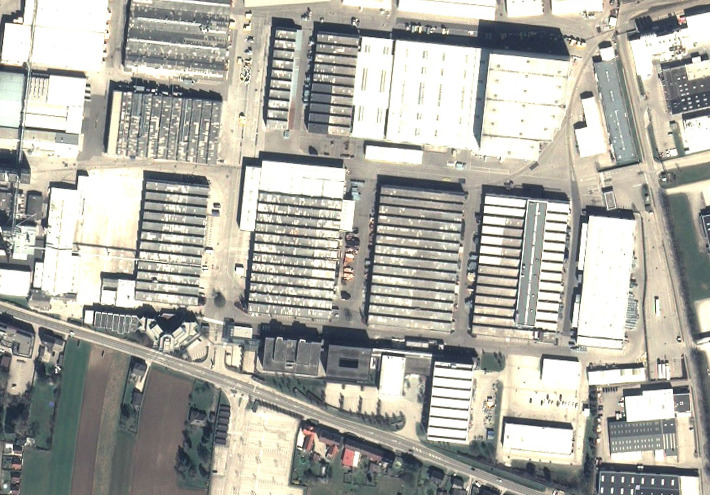}}}&
\raisebox{-.5\height}{\frame{\includegraphics[width=0.16\linewidth]{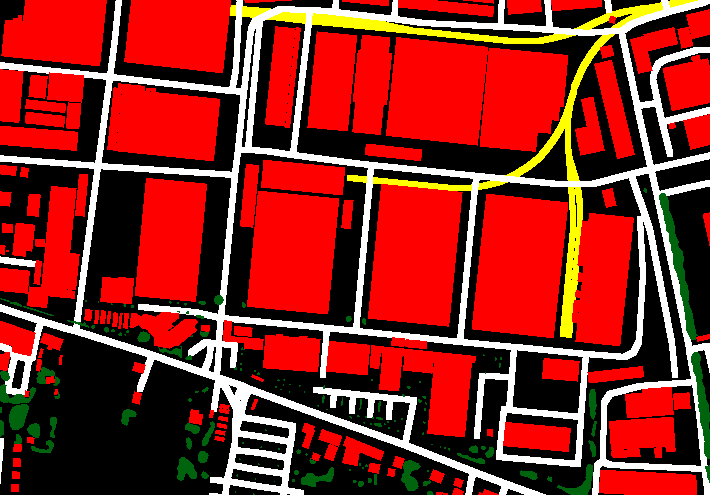}}}&
\raisebox{-.5\height}{\frame{\includegraphics[width=0.16\linewidth]{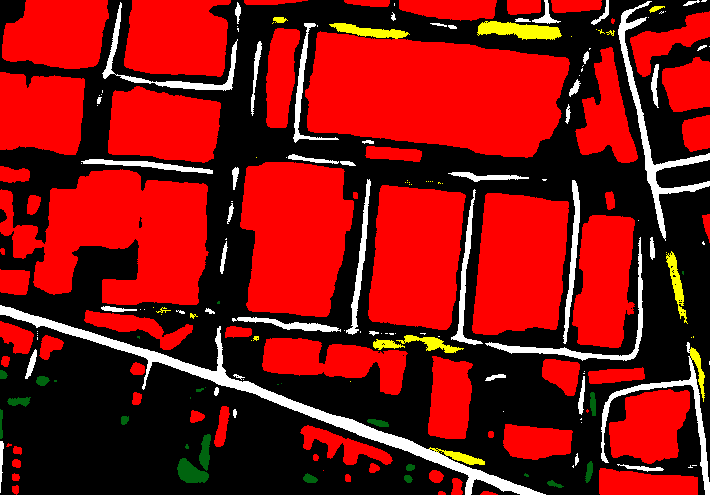}}}&
\raisebox{-.5\height}{\frame{\includegraphics[width=0.16\linewidth]{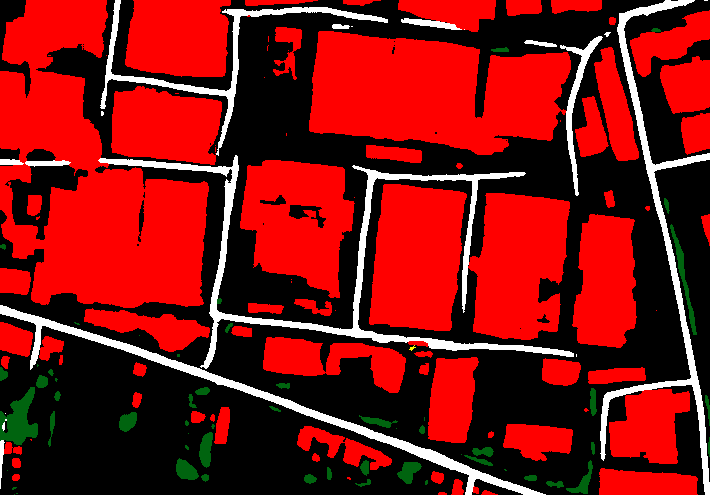}}}&
\raisebox{-.5\height}{\frame{\includegraphics[width=0.16\linewidth]{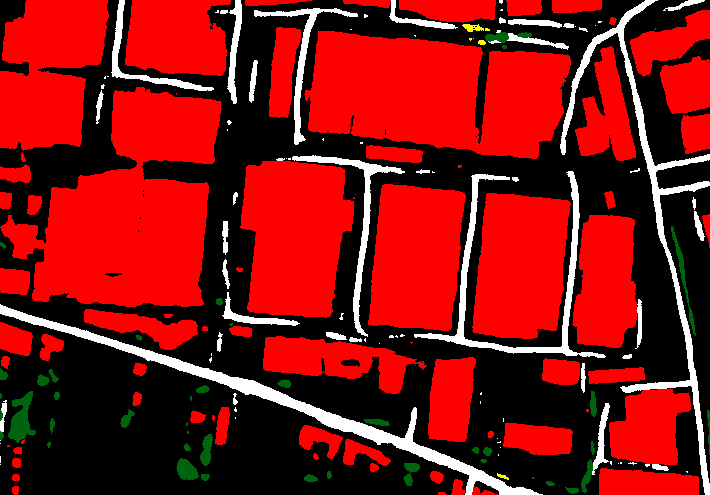}}}&
\raisebox{-.5\height}{\frame{\includegraphics[width=0.16\linewidth]{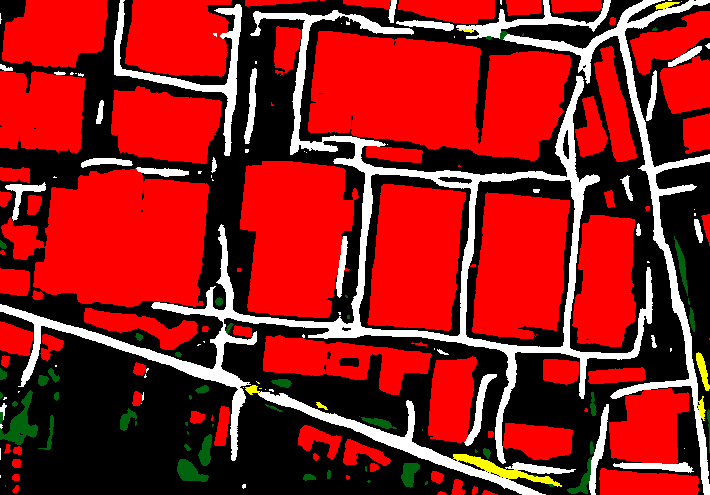}}}
\\[2.67em]
\rotatebox[origin=c]{90}{\textit{Leibnitz}} &
\raisebox{-.5\height}{\frame{\includegraphics[width=0.16\linewidth]{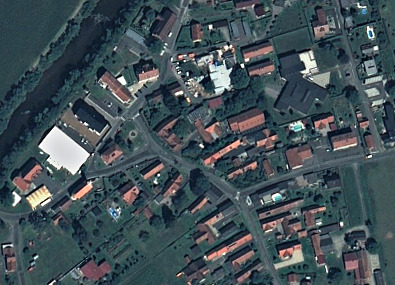}}}&
\raisebox{-.5\height}{\frame{\includegraphics[width=0.16\linewidth]{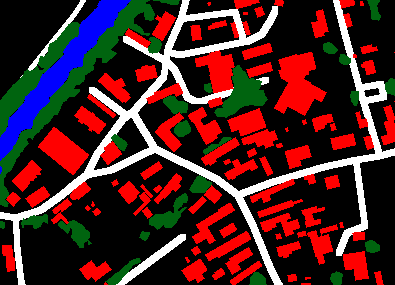}}}&
\raisebox{-.5\height}{\frame{\includegraphics[width=0.16\linewidth]{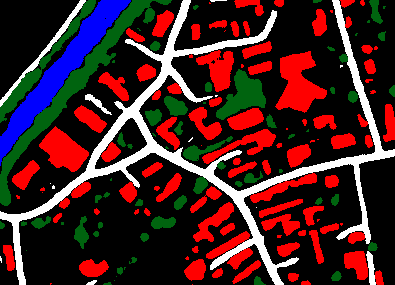}}}&
\raisebox{-.5\height}{\frame{\includegraphics[width=0.16\linewidth]{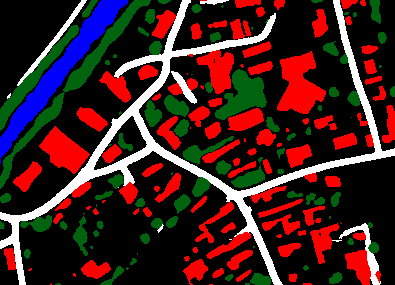}}}&
\raisebox{-.5\height}{\frame{\includegraphics[width=0.16\linewidth]{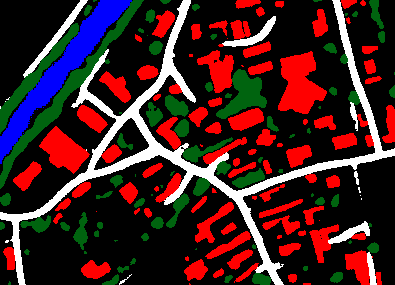}}}&
\raisebox{-.5\height}{\frame{\includegraphics[width=0.16\linewidth]{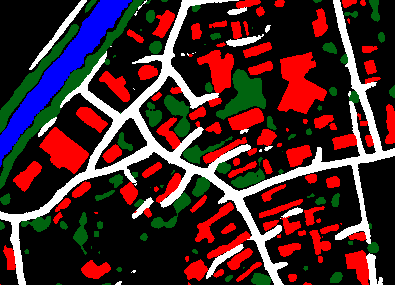}}}
\\[2.8em]
\rotatebox[origin=c]{90}{\textit{Lille}} &
\raisebox{-.5\height}{\frame{\includegraphics[width=0.16\linewidth]{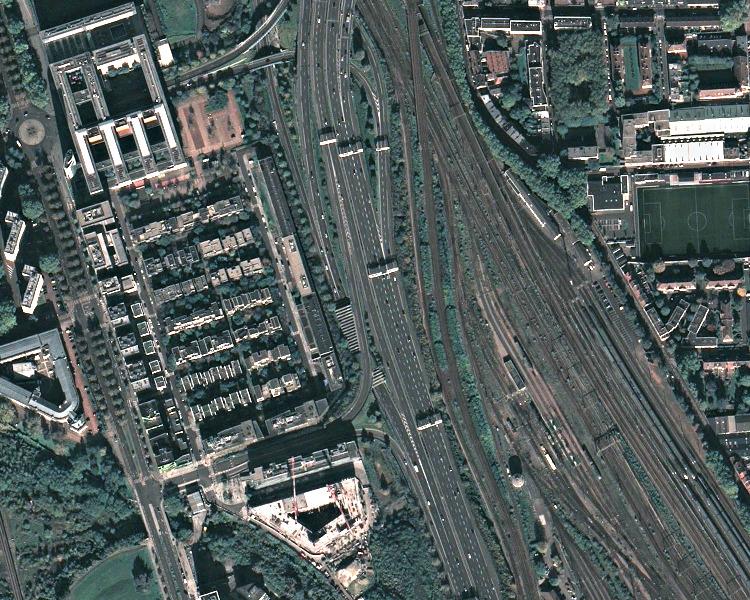}}}&
\raisebox{-.5\height}{\frame{\includegraphics[width=0.16\linewidth]{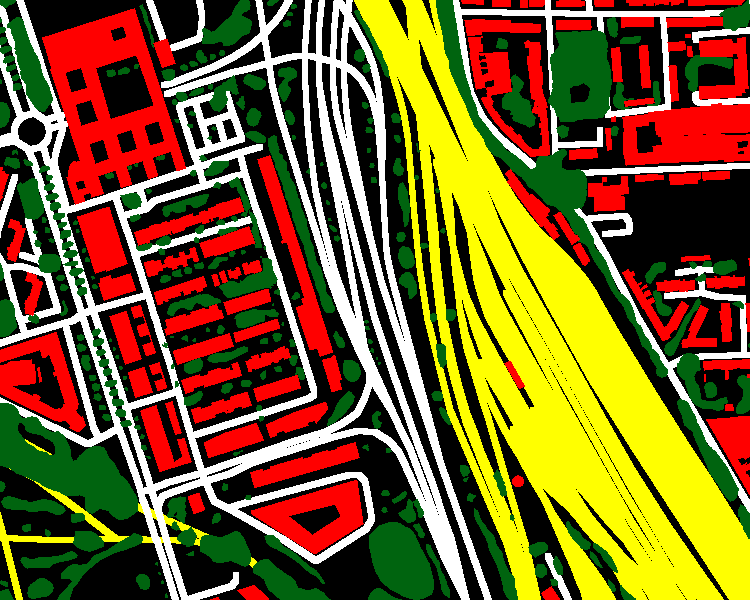}}}&
\raisebox{-.5\height}{\frame{\includegraphics[width=0.16\linewidth]{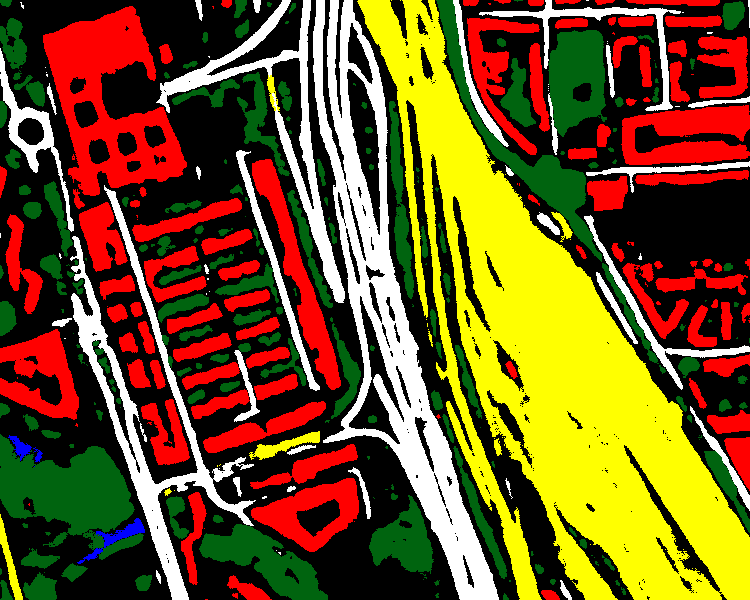}}}&
\raisebox{-.5\height}{\frame{\includegraphics[width=0.16\linewidth]{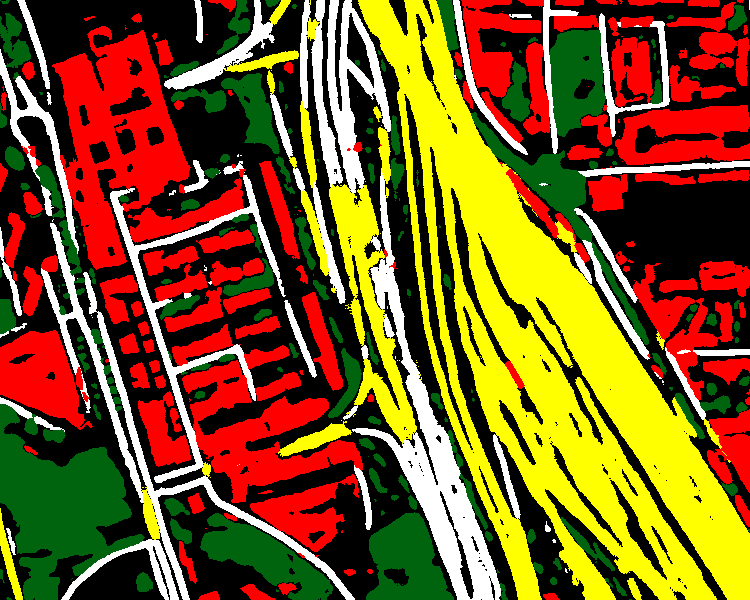}}}&
\raisebox{-.5\height}{\frame{\includegraphics[width=0.16\linewidth]{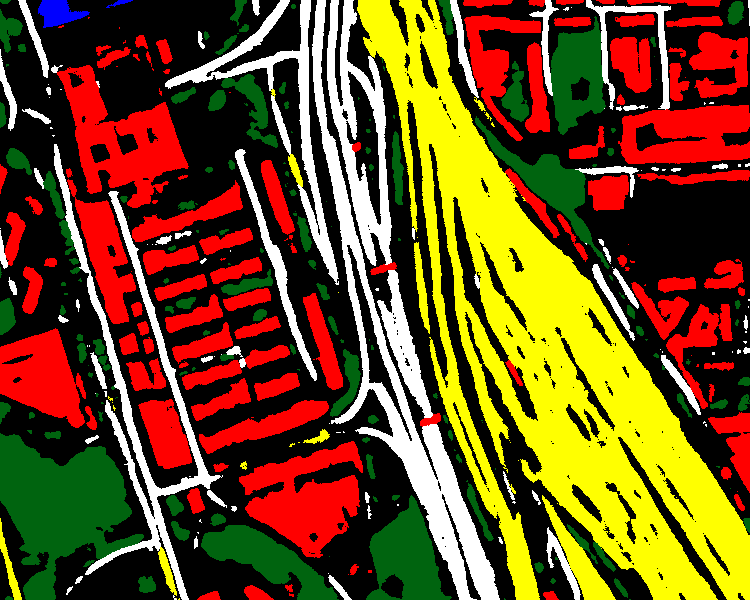}}}&
\raisebox{-.5\height}{\frame{\includegraphics[width=0.16\linewidth]{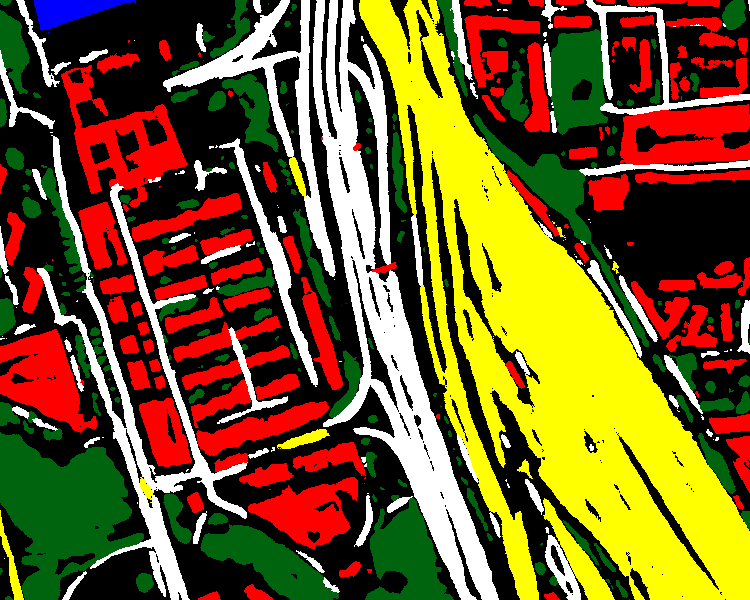}}}
\\[3.1em]
\rotatebox[origin=c]{90}{\textit{Roanne}} &
\raisebox{-.5\height}{\frame{\includegraphics[width=0.16\linewidth]{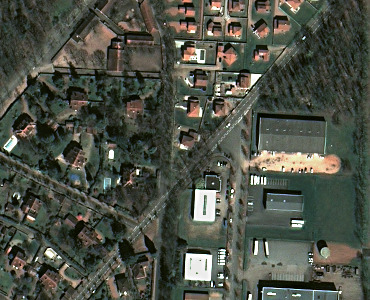}}}&
\raisebox{-.5\height}{\frame{\includegraphics[width=0.16\linewidth]{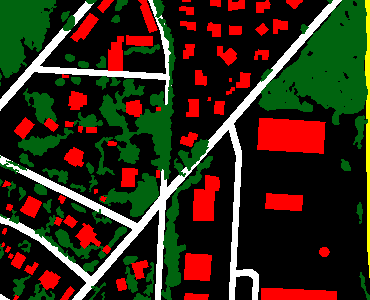}}}&
\raisebox{-.5\height}{\frame{\includegraphics[width=0.16\linewidth]{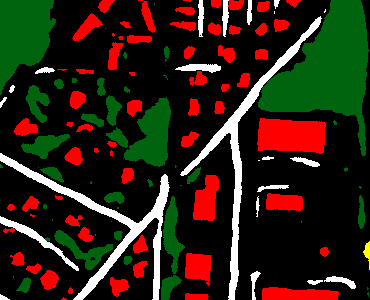}}}&
\raisebox{-.5\height}{\frame{\includegraphics[width=0.16\linewidth]{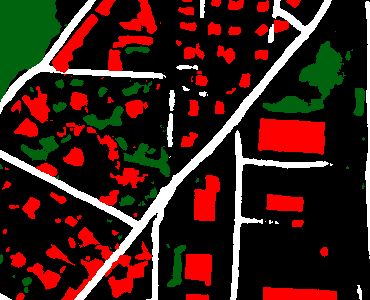}}}&
\raisebox{-.5\height}{\frame{\includegraphics[width=0.16\linewidth]{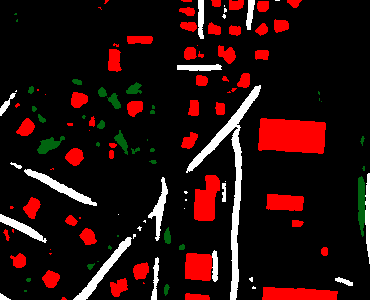}}}&
\raisebox{-.5\height}{\frame{\includegraphics[width=0.16\linewidth]{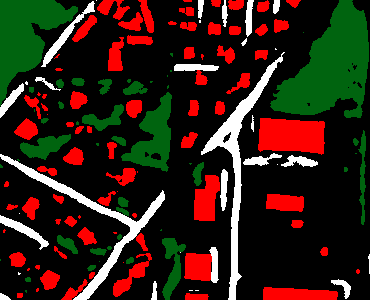}}}
\end{tabular}
\caption{Close-ups from validation cities of the Luxcarta dataset. Classes: background (black), building (red), road (white), railway (yellow), high vegetation (green), and water (blue).}
\label{fig:close_results_luxcarta}
\end{figure*}

The overall F1-scores of all the classes on the Luxcarta dataset for each method are reported in Table~\ref{table:luxcarta_f1}. The method, which achieves the most similar performance with \textit{static learning} is highlighted. Fig.~\ref{fig:luxcarta_plots} depicts the change of IoU values on the validation cities as the training progresses. Visual close-up results for \textit{static learning}, \textit{multiple learning}, \textit{incremental learning w/o rem} and \textit{incremental learning} generated by the final models are shown in Fig.~\ref{fig:close_results_luxcarta}. Although our network generates a binary label map for each class, for the sake of compact and better visualization, we provide multi-class predicted maps obtained by assigning each pixel to the class, for which the highest probability is produced. In the figure, the pixels, having no probability higher than or equal to 0.5 are labeled as background. 

As expected, \textit{static learning}, outperforms the other approaches on the Luxcarta dataset (see Table~\ref{table:luxcarta_f1}), because in the training stage, we feed much more and diverse training data to the model compared to the other approaches. Although \textit{static learning} is superior to the other approaches on the Luxcarta dataset, it is applicable only if the data are static and the annotations are unique, which is almost never the case in real-world applications. In \textit{multiple learning}, even if the previous data are not accessible, predicted maps for all the presented classes can be generated. However, because of the growing number of classifiers, this approach is inefficient in terms of test efficiency and storage. In addition, for each individual classifier, learning is limited to the data, on which the classifier was initially trained. For instance, \textit{building - high vegetation} classifier trained on Train1 can not be fine-tuned on Train2, as annotations for these classes are not available on Train2.

In \textit{fixed representation} methodology, the exact performance for the initially introduced classes is retained as neither the shared nor the classification layers for these classes change. On the other hand, the network performs extremely poorly for the new classes as shown in Fig. \ref{fig:luxcarta_plots_fe} and reported in Table~\ref{table:luxcarta_f1}. All in all, we conclude that shared layers of the network have to be adapted to the new training data. When we apply \textit{fine-tuning}, since instead of initializing all the parameters randomly, the extracted features for the previous classes are used, performance for the new classes is remarkable, especially when there is only one class to be added. For instance, it is the best performer for \textit{water} class. However, the results justify that the network catastrophically forgets the previously learned information.

\begin{figure}
	\centering
	\includegraphics[width=0.99\linewidth]{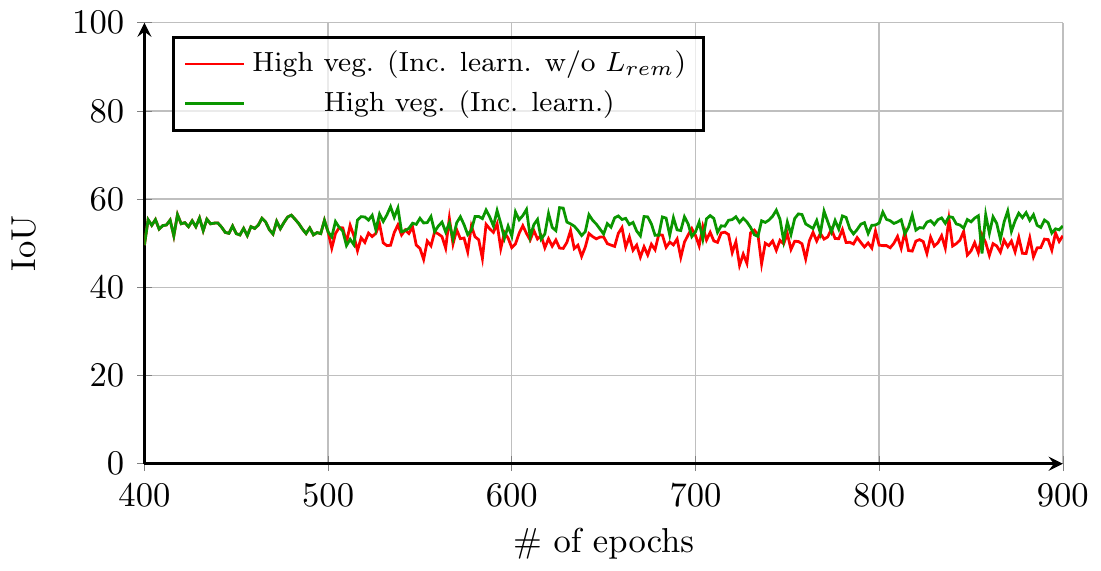}
  	\caption{Comparison of \textit{incremental learning} and \textit{incremental learning w/o $L_{rem}$} for \textit{high vegetation} class.}
	\label{fig:inc_wo_and_w_old_data}
\end{figure}

As reported in Table~\ref{table:luxcarta_f1}, \textit{incremental learning} exhibits the closest performance to \textit{static learning}. Since our approach enables the network to learn the old classes on the new data and remember them from the previous data, performance for the previous classes gets better over time. If the previous data are never shown, performance for the old classes may decrease as a result of adapting the network to the new data completely and imprecision of output of the memory network on the new data for the previous classes. Fig.~\ref{fig:inc_wo_and_w_old_data} compares \textit{incremental learning} and \textit{incremental learning w/o $L_{rem}$} for \textit{high vegetation} before and after adding \textit{road} and \textit{railway} classes on Train2 (i.e., before and after the $500^{th}$ epoch) to the \textit{building} $\&$ \textit{high vegetation} classifier trained on Train1. The close-ups from \textit{Roanne} in Fig.~\ref{fig:close_results_luxcarta} show that \textit{incremental learning w/o $L_{rem}$} fails in detecting a lot of \textit{high vegetation}, whereas \textit{incremental learning} exhibits a good performance. We also observe that \textit{incremental learning} significantly outperforms \textit{multiple learning} for \textit{building} class. The reason is that the network in \textit{multiple learning} learns \textit{building} only on Train1, while \textit{incremental learning} facilitates learning the same class from all the training data sequentially. Although when buildings are small and regular shaped as in \textit{Leibnitz} and \textit{Roanne}, both approaches generate similar outputs, \textit{multiple learning} is not able to delineate the borders very well when buildings cover a large area as in \textit{Amstetten}. \textit{Road} and \textit{Railway} classes turn out to be the most difficult classes, as the numeric results for them are much lower than the others. As can be seen in the close-up from \textit{Lille}, they visually look quite similar, which makes the classifiers confuse between them in some cases. \textit{Incremental learning} seems detecting the roads and railways that are mis-classified by \textit{incremental learning w/o $L_{rem}$}.

\subsection{Experiments on the Benchmark Datasets}\label{sec:experiments_on_benchmarks}
In the experiments on the benchmarks, we assume that we have access the whole training tiles, but we are provided the annotations sequentially. We suppose that every time a new set of annotations are retrieved, the previous one is not accessible. On the Vaihingen dataset, we consider that we retrieve label maps for \textit{building} and \textit{high vegetation} classes in the beginning. We are then given the ground-truth for \textit{impervious surfaces} and \textit{low vegetation}. Finally, we receive the annotations for \textit{car} class. On the Potsdam dataset, since there is an additional \textit{clutter} class, we assume that the label map for this class is also available in the initial training data. For our approach, since we always use the same training images, we remind the network the old classes using output of the memory network (i.e., we only optimize $L_{adapt}$). On contrary the other approaches, for \textit{static learning}, we use all training tiles as well as annotations for all the classes at once in the training stage.

\begin{table*}
\centering
\caption{F1 scores on the Vaihingen benchmark dataset.}
\label{table:vaihingen_f1}
\begin{tabular}{|c|c|ccccc||c||}
\hline
\multicolumn{1}{|c}{} & \textbf{Epoch} & \multicolumn{1}{||c}{\textbf{Building}} & \textbf{High veg.} & \textbf{Imper. surf.} & \textbf{Low veg.} & \textbf{Car} & \textbf{Overall}\\
\hline
\textit{static learning}
 & \textbf{500}   & \multicolumn{1}{||c}{93.61 (Ref.)} & 87.87 (Ref.)              & 91.55 (Ref.)           & 81.05 (Ref.)             & 82.83 (Ref.)   &   87.38 (Ref.)    \\
\hline 
\hline
\textit{multiple learning} 
 & \textbf{500}   & \multicolumn{1}{||c}{\textbf{94.43}} & \textbf{88.12}  & \multicolumn{1}{||c}{90.71}          & \multicolumn{1}{c||}{80.41}  & \textbf{87.90} &  \textbf{88.31 (+0.93)} \\
\hline
\multirow{2}{*}{\textit{fixed representation}}
 & 1000  & \multicolumn{1}{||c}{94.43} & 88.12              & \multicolumn{1}{||c}{87.09}          & \multicolumn{1}{c||}{76.39} & \textemdash  &  \\
 & \textbf{1500}  & \multicolumn{1}{||c}{\textbf{94.43}} & \textbf{88.12}     & \multicolumn{1}{||c}{87.09}          & \multicolumn{1}{c||}{76.39} & 13.37   &   71.88 (-15.5)    \\
\hline
\multirow{2}{*}{\textit{fine-tuning}}
 & 1000  & \multicolumn{1}{||c}{52.40} & 0.03               & \multicolumn{1}{||c}{91.83} & \multicolumn{1}{c||}{80.99} & \textemdash  &  \\
 & \textbf{1500}  & \multicolumn{1}{||c}{0.02} & 0.00               & \multicolumn{1}{||c}{43.81}          & \multicolumn{1}{c||}{0.01}  & 86.18        & 26.00 (-61.38) \\
\hline  
\textit{incremental learning}
& 1000 & \multicolumn{1}{||c}{94.34} & 88.02              & \multicolumn{1}{||c}{91.42}          & \multicolumn{1}{c||}{81.65} & \textemdash &   \\
\textit{w/o $L_{rem}$}
 & \textbf{1500}  & \multicolumn{1}{||c}{94.31} & 88.07              & \multicolumn{1}{||c}{\textbf{91.51}}          & \multicolumn{1}{c||}{\textbf{81.60}}          & 81.69      & 87.44 (+0.06)  \\
\hline
\multicolumn{1}{c}{} &  & \multicolumn{2}{||c|}{\textbf{Training Set 1}}  & \multicolumn{2}{||c||}{\textbf{Training Set 2}} & \textbf{Training Set 3} \\
\cline{3-7}  
\end{tabular}
\end{table*}

\begin{table*}
\centering
\caption{F1 scores on the Postdam benchmark dataset.}
\label{table:potsdam_f1}
\begin{tabular}{|c|ccccccc||c||}
\hline
\multicolumn{1}{|c}{} & \textbf{Epoch} & \multicolumn{1}{||c}{\textbf{Building}} & \textbf{High veg.}  & \textbf{Clutter} & \textbf{Imper. surf.} &  \textbf{Low veg.} & \textbf{Car} & \textbf{Overall}  \\
\hline
\textit{static learning}
 & \textbf{500}   & \multicolumn{1}{||c}{96.83 (Ref.)} & 85.04 (Ref.)  & 54.57 (Ref.) & 92.62 (Ref.)    & 85.69 (Ref.) & 94.84 (Ref.) &  84.93 (Ref.) \\
\hline
\hline
\textit{multiple learning}
 & \textbf{500}   & \multicolumn{1}{||c}{96.59} & 85.25 & \multicolumn{1}{c||}{50.82} & 92.07 & \multicolumn{1}{c||}{84.82} & \textbf{95.36} & 84.15 (-0.78) \\
\hline
\multirow{2}{*}{\textit{fixed representation}}
 & 1000  & \multicolumn{1}{||c}{96.59} & 85.25 & \multicolumn{1}{c||}{50.82} & 86.76 & \multicolumn{1}{c||}{79.98} & \textemdash & \\
 & \textbf{1500}  & \multicolumn{1}{||c}{96.59} & 85.25 & \multicolumn{1}{c||}{50.82} & 86.76 & \multicolumn{1}{c||}{79.98} & 72.14 &      78.59 (-6.34) \\
\hline
\multirow{2}{*}{\textit{fine-tuning}}
 & 1000  & \multicolumn{1}{||c}{0.00} & 44.53 & \multicolumn{1}{c||}{3.23} & 92.13 & \multicolumn{1}{c||}{85.45} & \textemdash & \\
 & \textbf{1500}  & \multicolumn{1}{||c}{1.62} & 24.73 & \multicolumn{1}{c||}{0.00} & 65.00 & \multicolumn{1}{c||}{0.01} & 94.60 &  30.99 (-54.94)     \\
\hline  
\textit{incremental learning}
 & 1000  & \multicolumn{1}{||c}{96.91} & 86.12 & \multicolumn{1}{c||}{50.23} & 92.20 & \multicolumn{1}{c||}{85.64} & \textemdash & \\
\textit{w/o $L_{rem}$}
 & \textbf{1500}  & \multicolumn{1}{||c}{\textbf{96.86}} & \textbf{85.28} & \multicolumn{1}{c||}{\textbf{51.56}}          & \textbf{92.10}  & \multicolumn{1}{c||}{\textbf{85.28}} & 94.43 & \textbf{84.25 (-0.68)}    \\
\hline
\multicolumn{1}{c}{} &  & \multicolumn{3}{||c||}{\textbf{Training Set 1}}  & \multicolumn{2}{|c||}{\textbf{Training Set 2}} & \textbf{Training Set 3} \\
\cline{3-8}  
\end{tabular}
\end{table*}

Because the images in the benchmarks are of much higher resolution than the satellite images in the Luxcarta dataset, the patches need to be larger to cover a reasonable area. Therefore, we divide the training tiles into 512 $\times$ 512 patches. The validation tiles are split into 2000 $\times$ 2000 patches. The training and validation tiles have 64 $\times$ 64 and 120 $\times$ 120 pixels of overlap, respectively. We compute a global mean for each channel from the training tiles and subtract it from all the pixels. 

For each approach, we train the same number of models for the same number of epochs and iterations using the same optimizer with the same parameters as in the experiments on the Luxcarta dataset. As size of the training patches is larger than in the previous experiments, we randomly sample 8 patches instead of 12. Another difference is that since both training and validation patches are from the same city, we augment the patches by only random flips and rotations. 

\begin{figure*}
\centering
\subfloat[Static learning\label{fig:vaihingen_plots_joint}]{\includegraphics[width=0.49\linewidth]{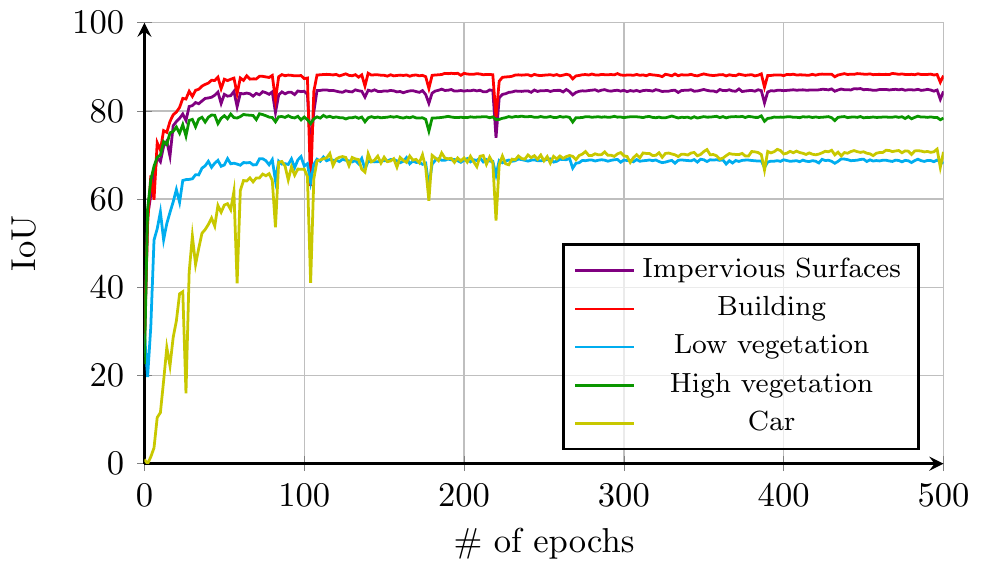}}
\hfill
\subfloat[Multiple learning\label{fig:vaihingen_plots_mult}]{\includegraphics[width=0.49\linewidth]{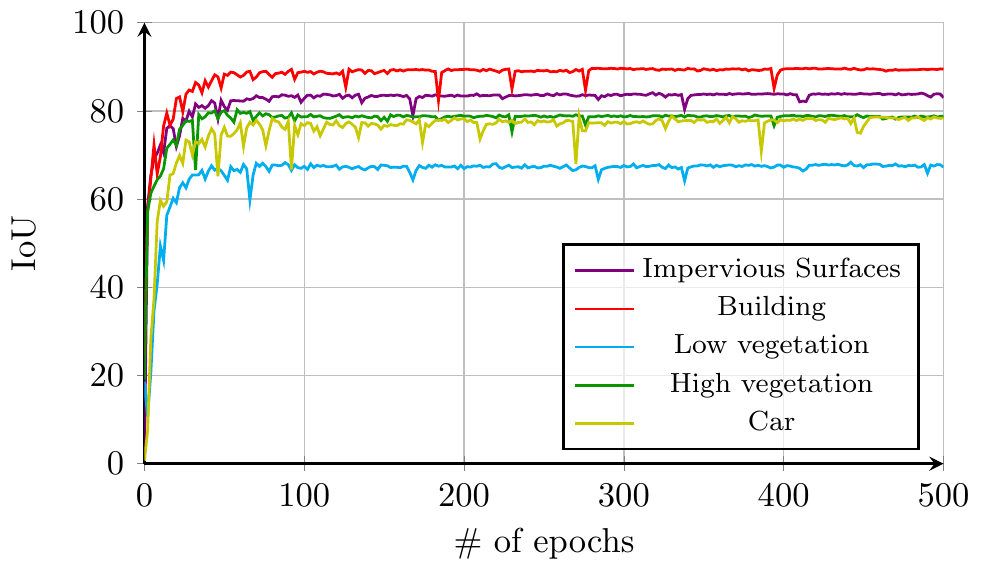}}
\hfill
\subfloat[Fixed representation\label{fig:vaihingen_plots_fe}]{\includegraphics[width=1\linewidth]{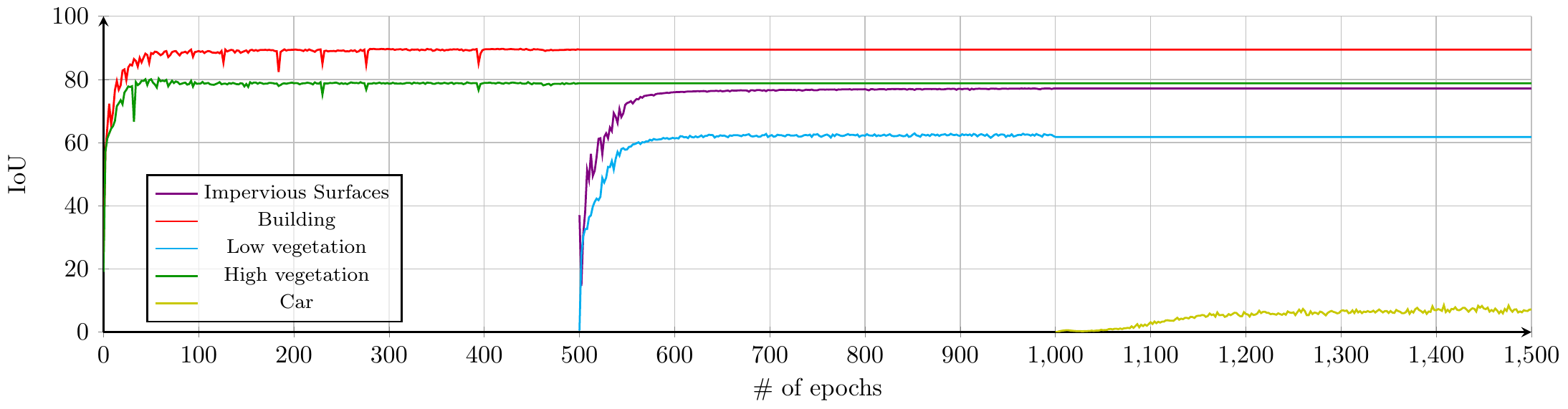}}
\hfill
\subfloat[Fine-tuning\label{fig:vaihingen_plots_ft}]{\includegraphics[width=1\linewidth]{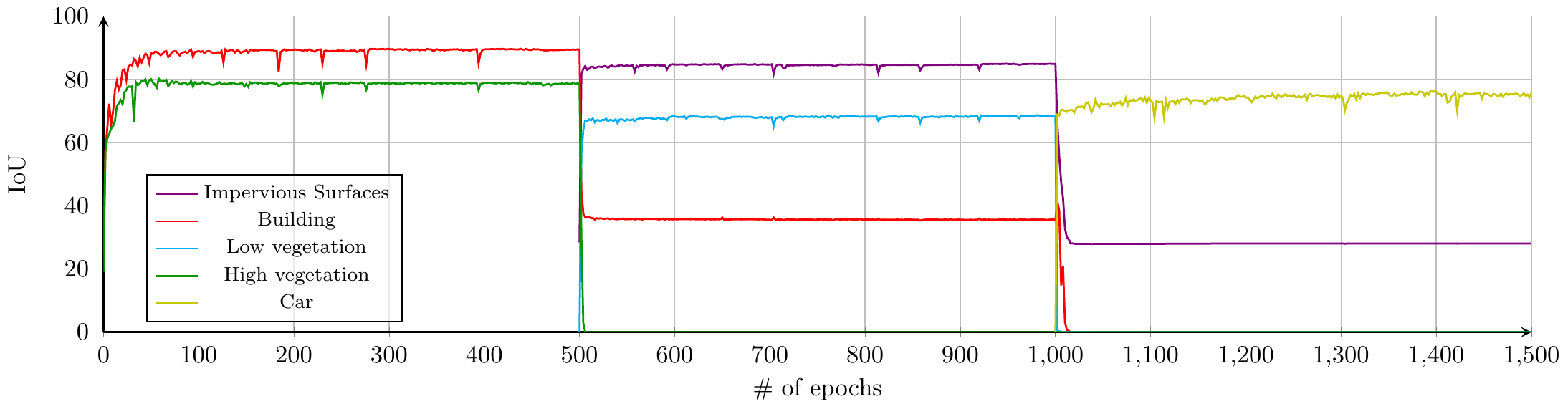}}
\hfill
\subfloat[Incremental learning w/o $L_{rem}$\label{fig:vaihingen_plots_inc}]{\includegraphics[width=1\linewidth]{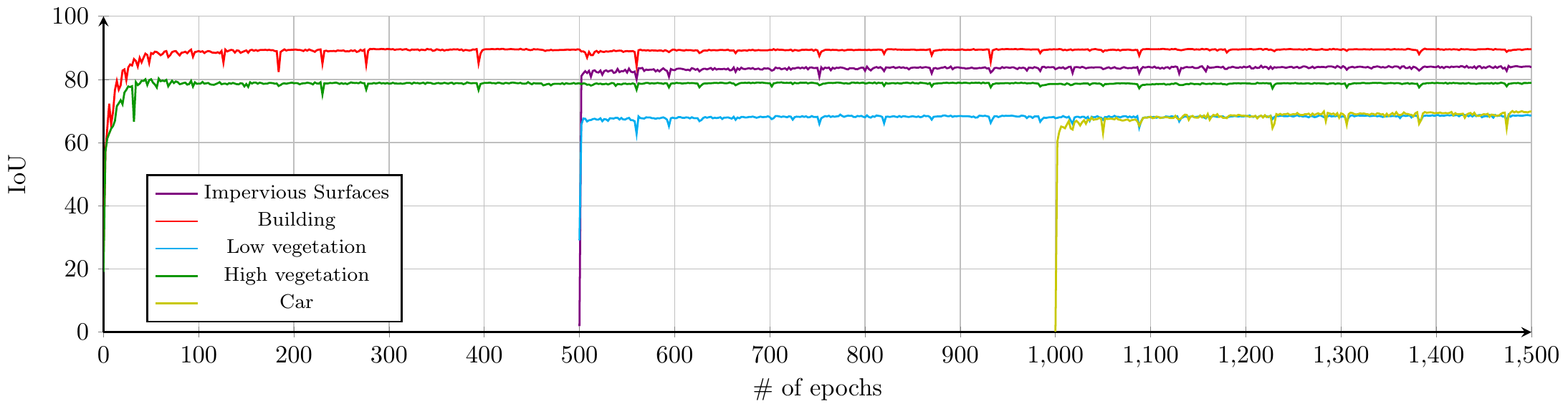}}
\hfill
\caption{Plots for the overall IoU values on validation data of the Vaihingen benchmark dataset.}
\label{fig:vaihingen_plots}
\end{figure*}

\begin{figure*}
\centering
\begin{tabular}{c@{\hspace{0.15em}}c@{\hspace{0.15em}}c@{\hspace{0.15em}}c@{\hspace{0.15em}}c}
\multirow{2}{*}{Image} & \multirow{2}{*}{Ground-truth} & \multirow{2}{*}{Static learning} & \multirow{2}{*}{Multiple learning} & Incremental learning \\
                       &                               &                                    &                                  & w/o $L_{rem}$ \\
\frame{\includegraphics[width=0.195\linewidth]{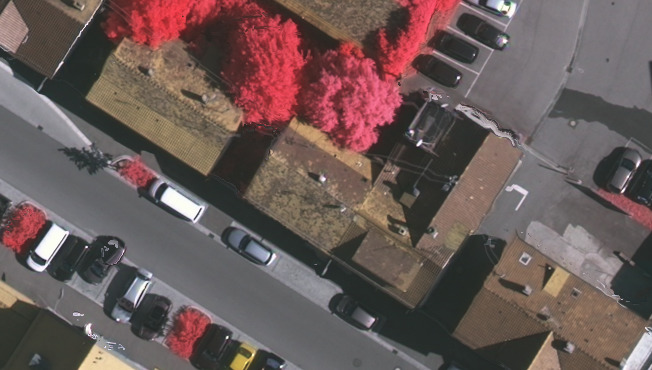}}&
\frame{\includegraphics[width=0.195\linewidth]{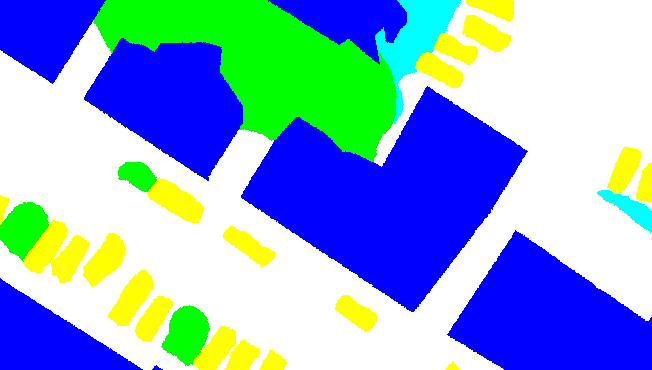}}&
\frame{\includegraphics[width=0.195\linewidth]{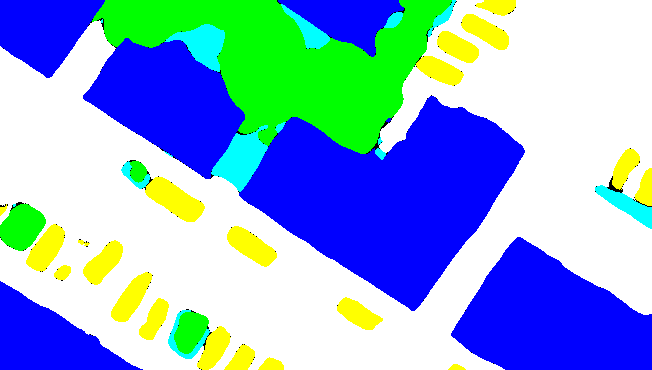}}&
\frame{\includegraphics[width=0.195\linewidth]{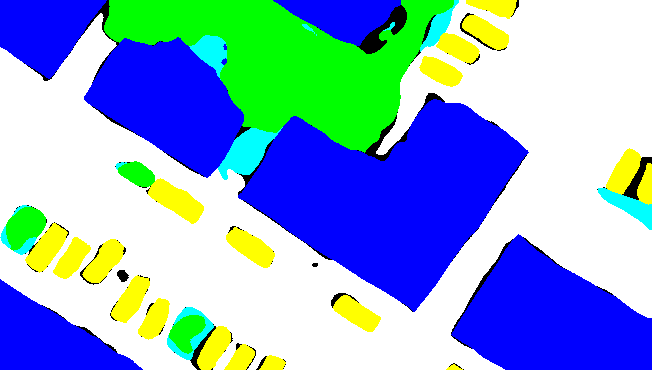}}&
\frame{\includegraphics[width=0.195\linewidth]{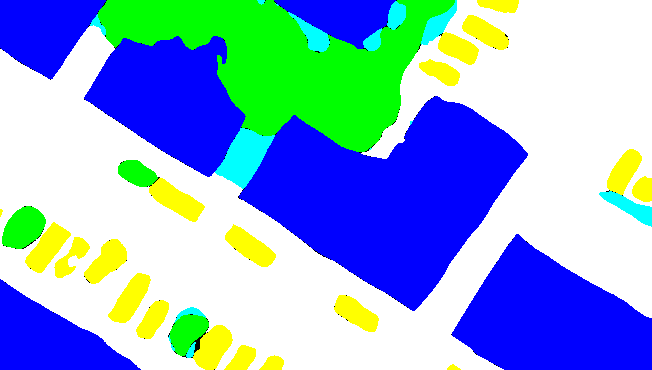}}\\[-0.2em]
\frame{\includegraphics[width=0.195\linewidth]{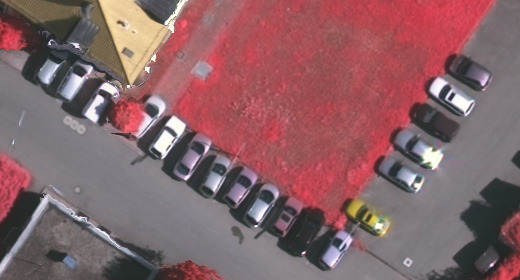}}&
\frame{\includegraphics[width=0.195\linewidth]{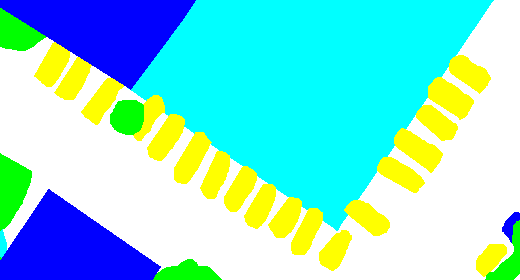}}&
\frame{\includegraphics[width=0.195\linewidth]{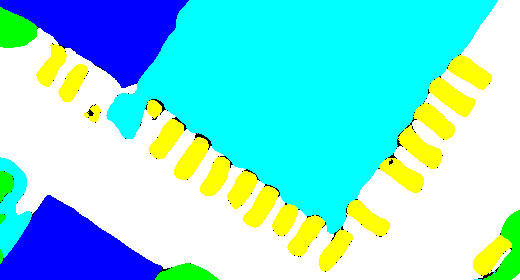}}&
\frame{\includegraphics[width=0.195\linewidth]{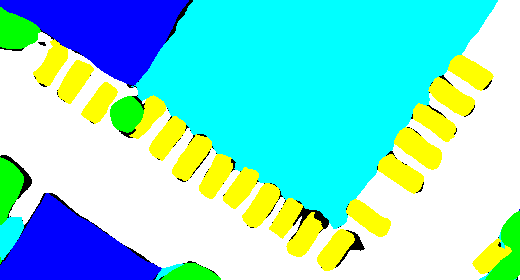}}&
\frame{\includegraphics[width=0.195\linewidth]{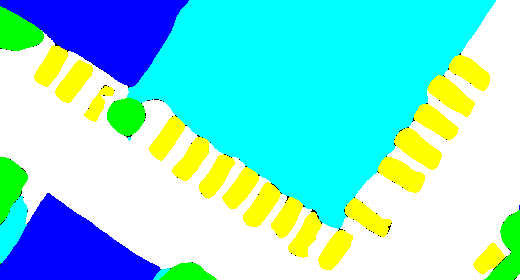}}\\[-0.2em]
\frame{\includegraphics[width=0.195\linewidth]{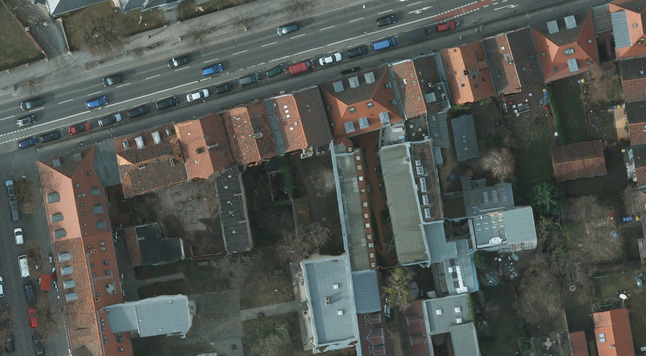}}&
\frame{\includegraphics[width=0.195\linewidth]{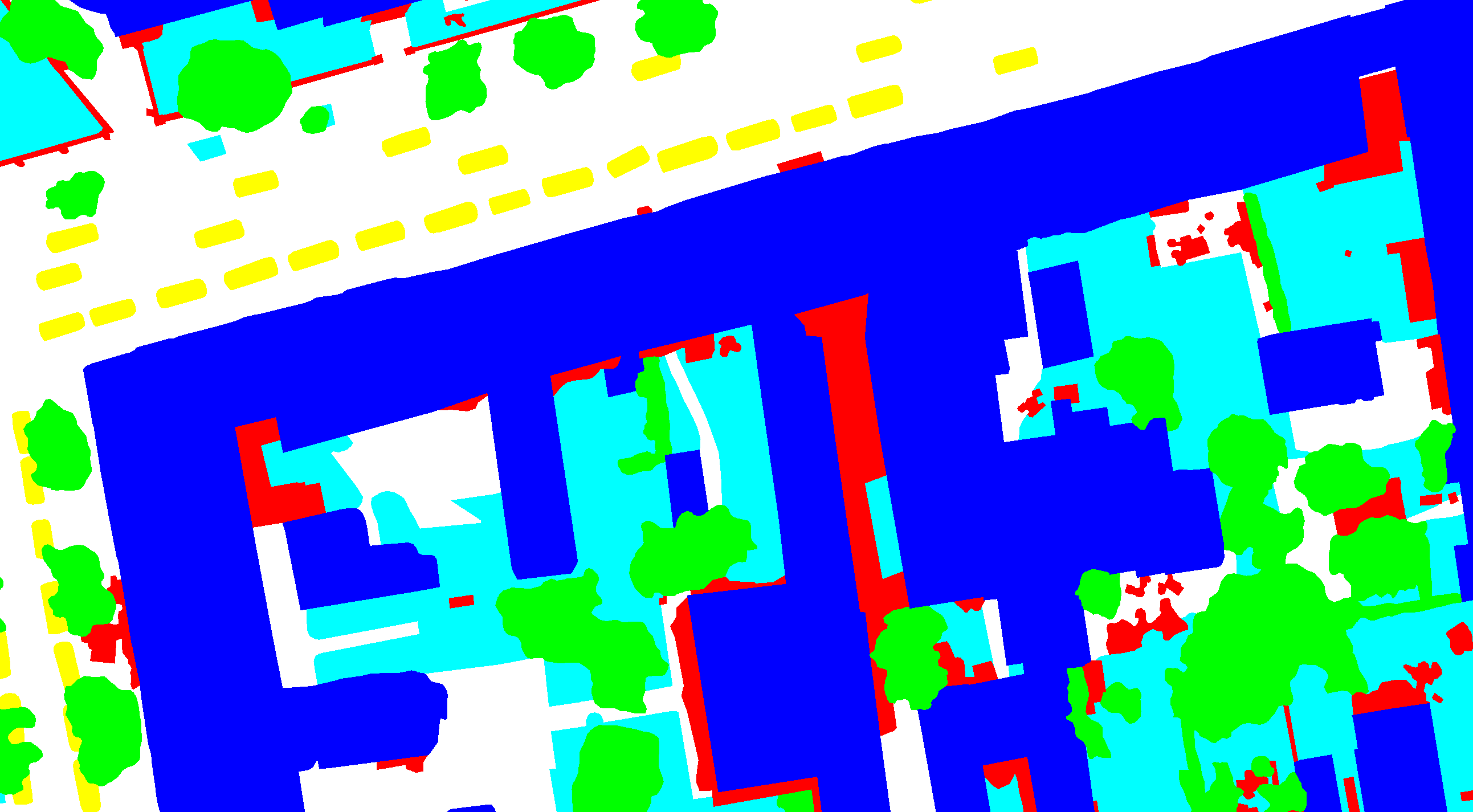}}&
\frame{\includegraphics[width=0.195\linewidth]{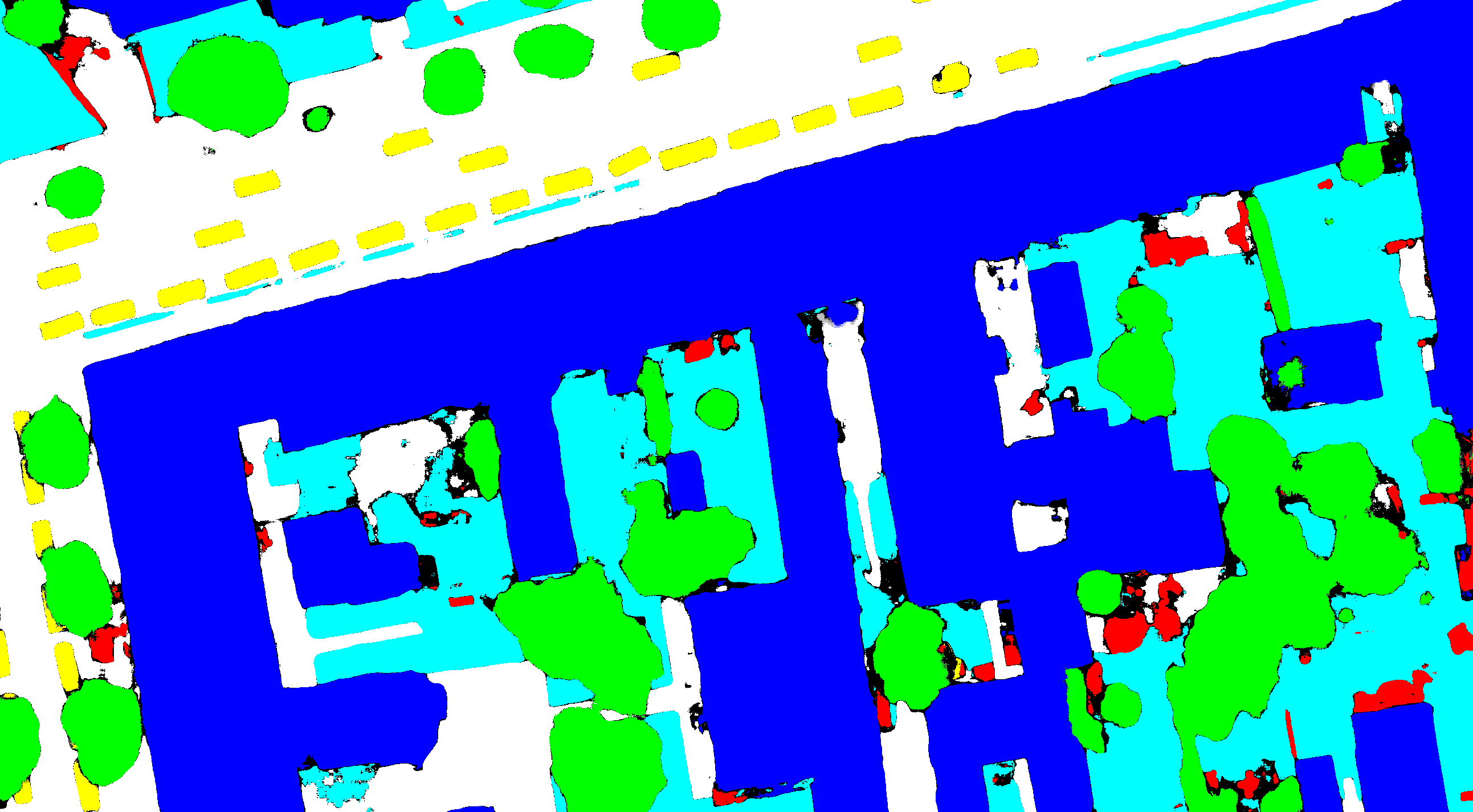}}&
\frame{\includegraphics[width=0.195\linewidth]{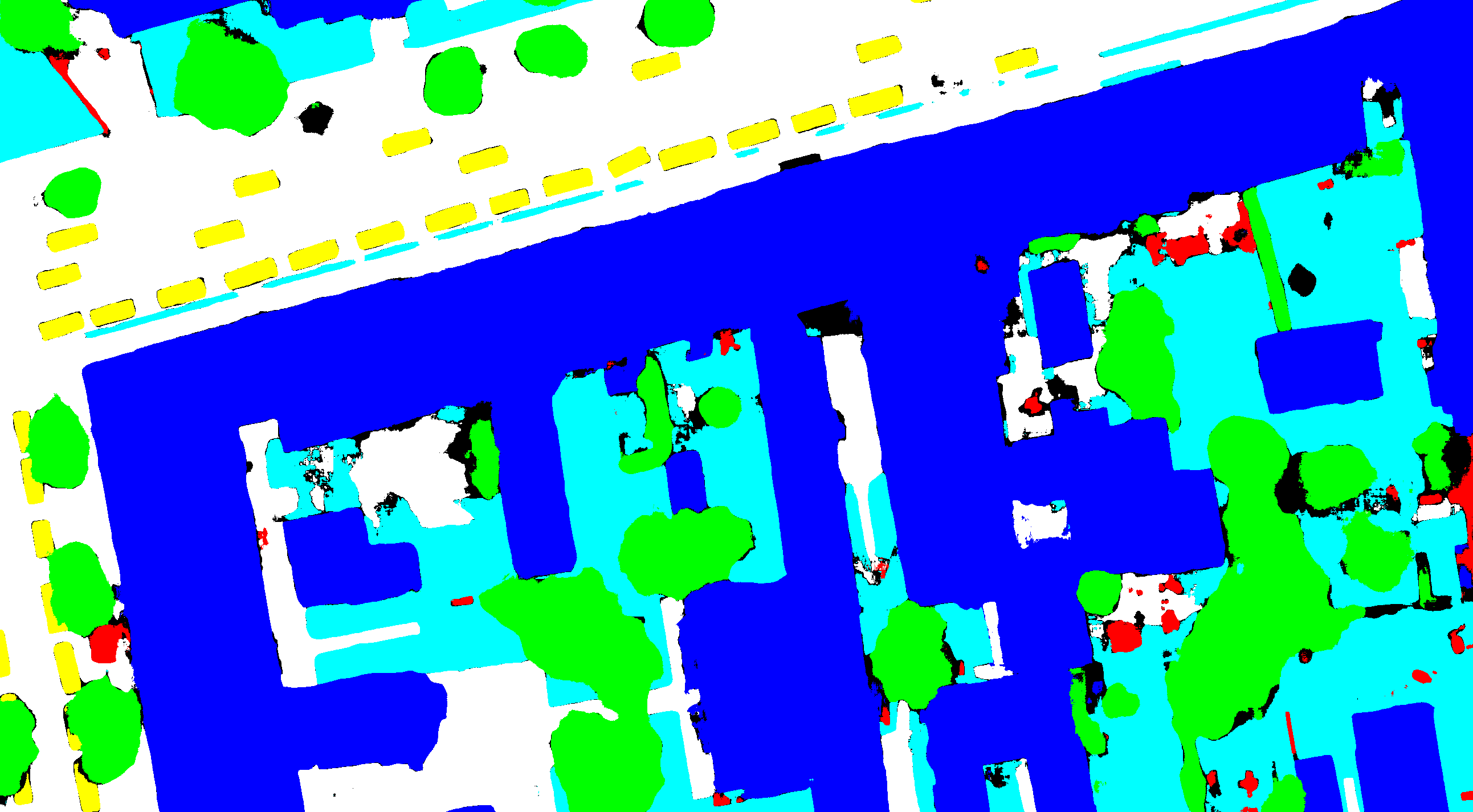}}&
\frame{\includegraphics[width=0.195\linewidth]{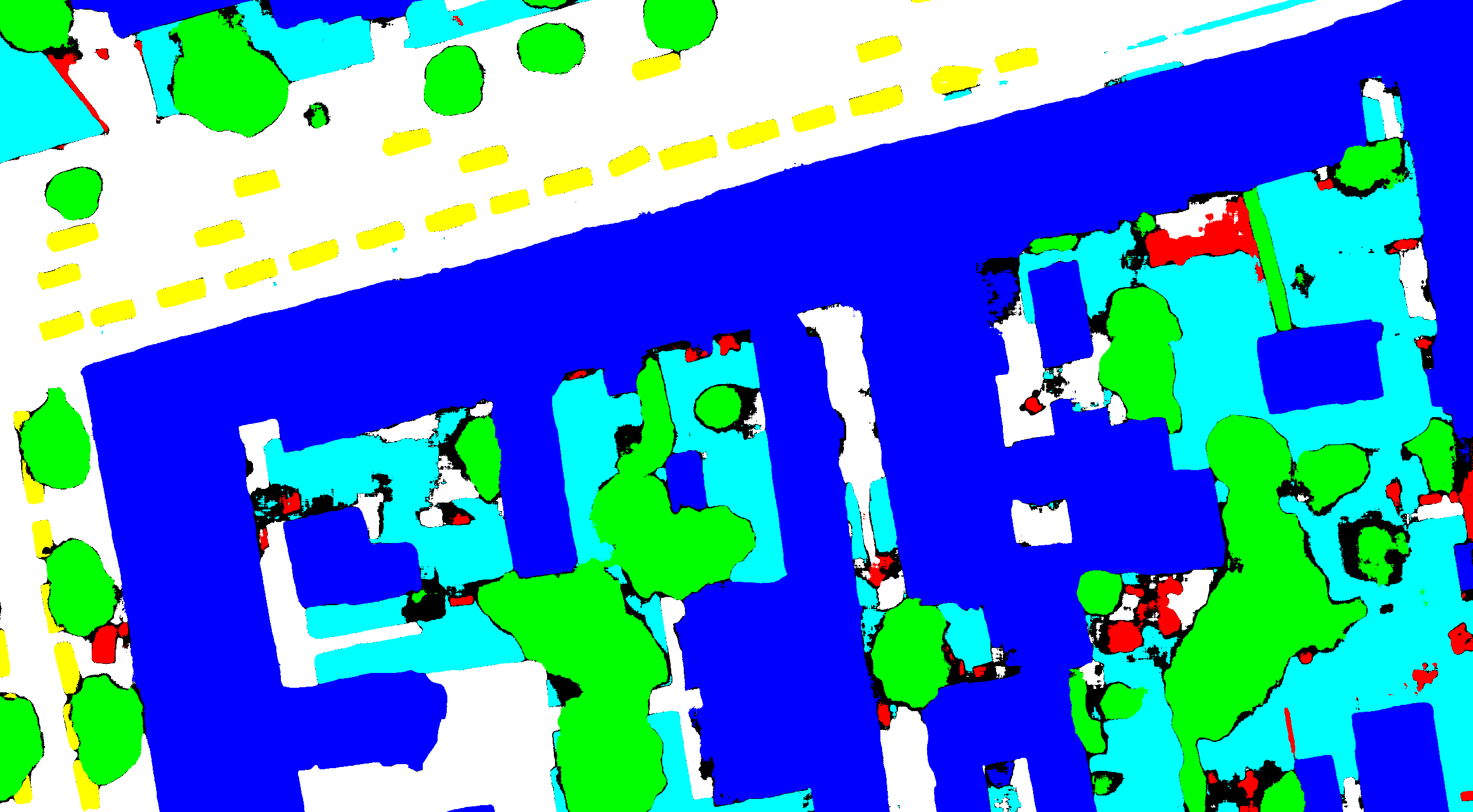}}\\[-0.2em]
\frame{\includegraphics[width=0.195\linewidth]{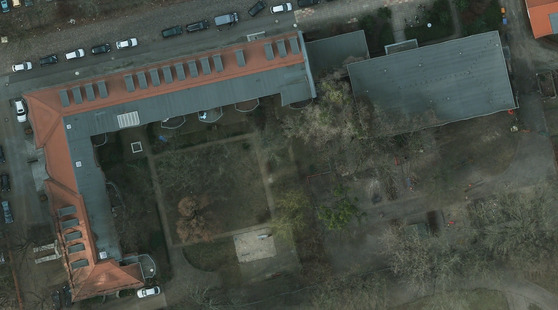}}&
\frame{\includegraphics[width=0.195\linewidth]{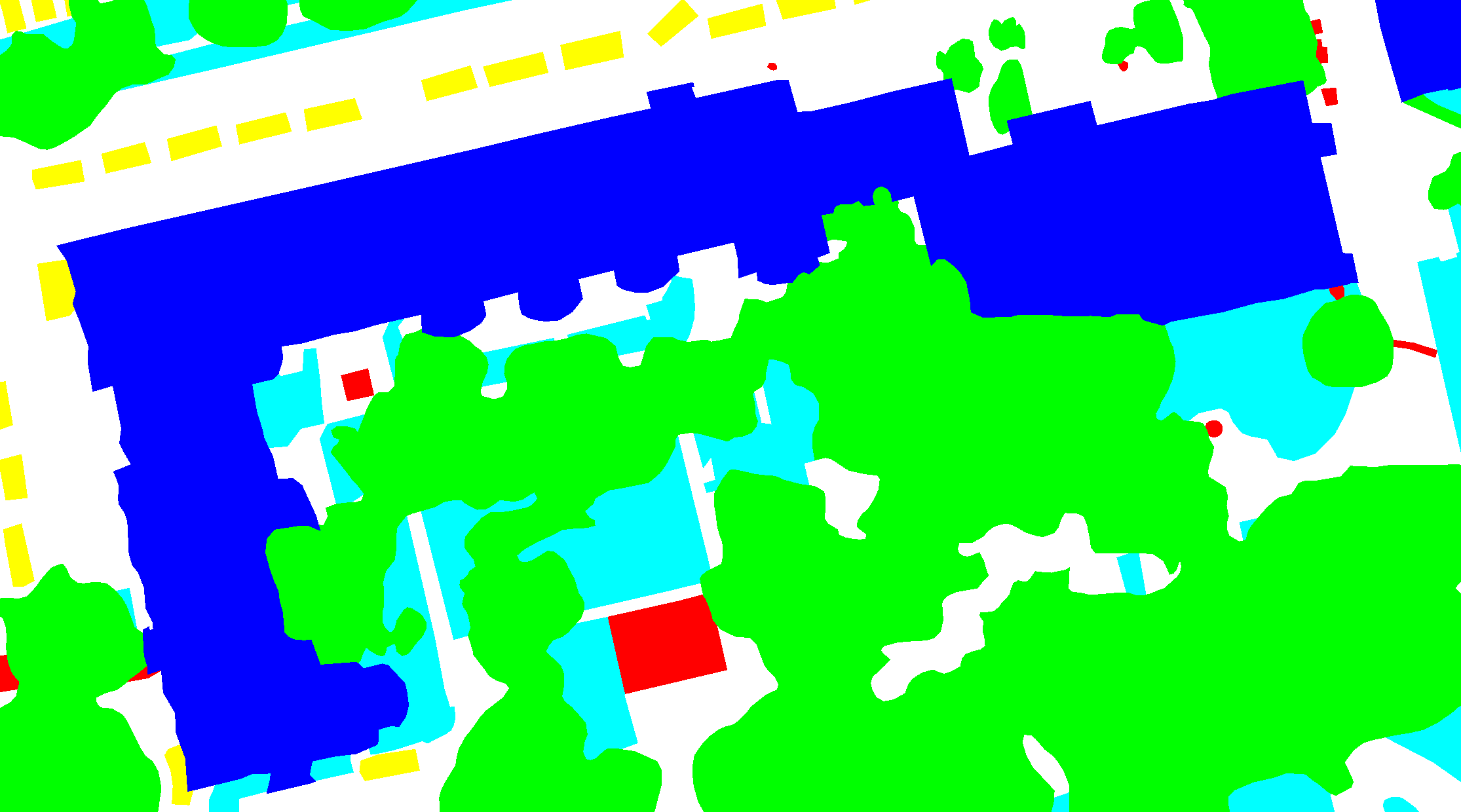}}&
\frame{\includegraphics[width=0.195\linewidth]{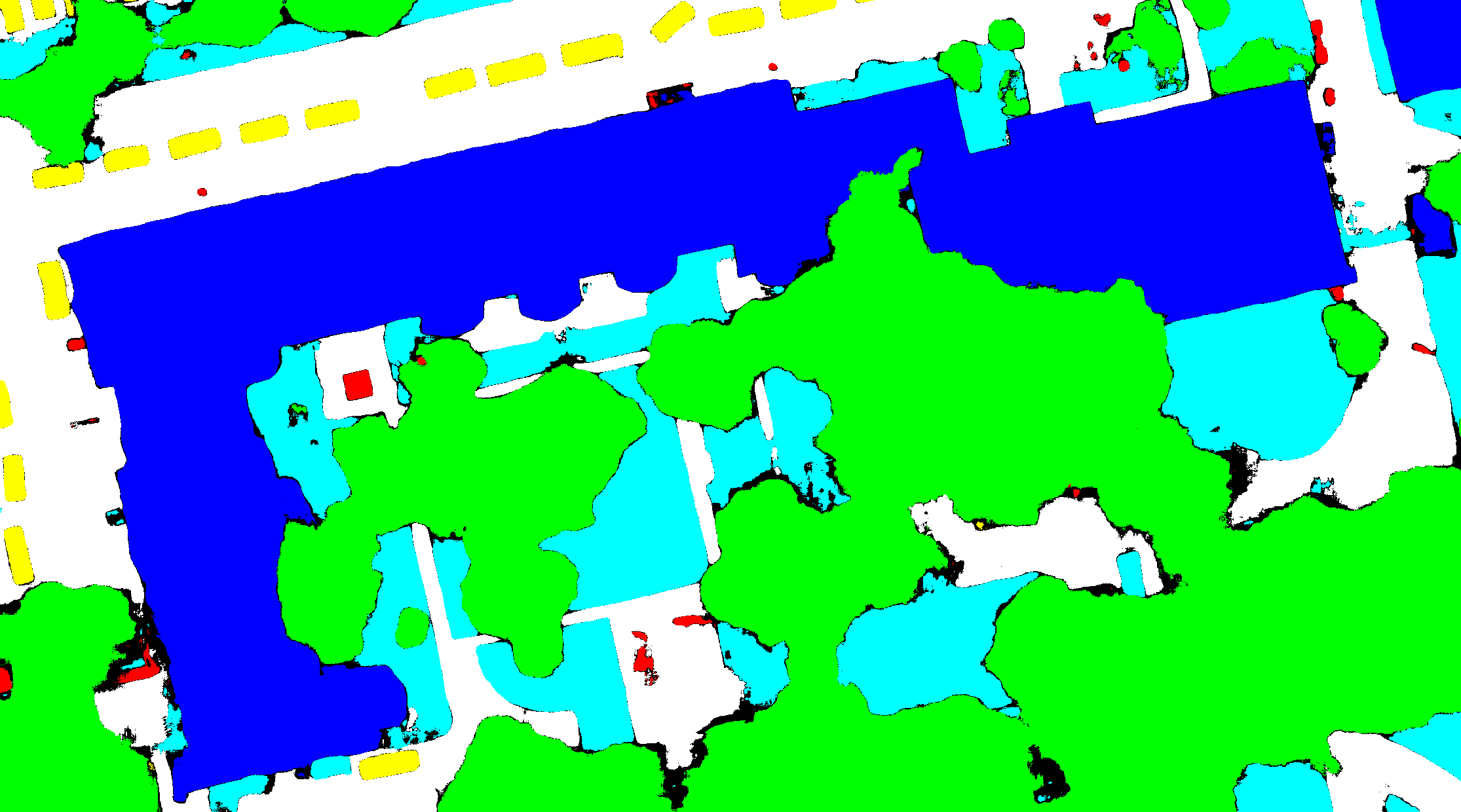}}&
\frame{\includegraphics[width=0.195\linewidth]{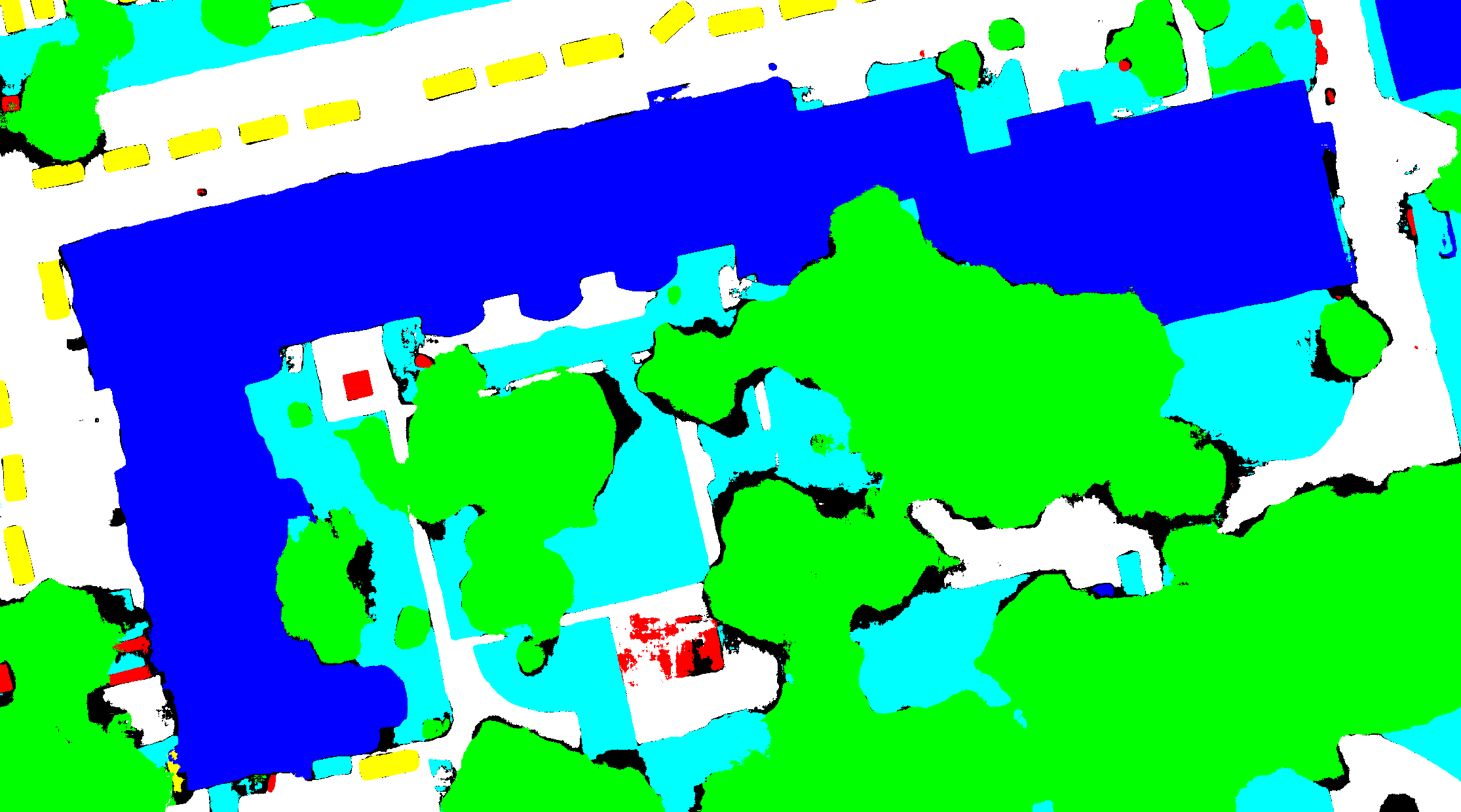}}&
\frame{\includegraphics[width=0.195\linewidth]{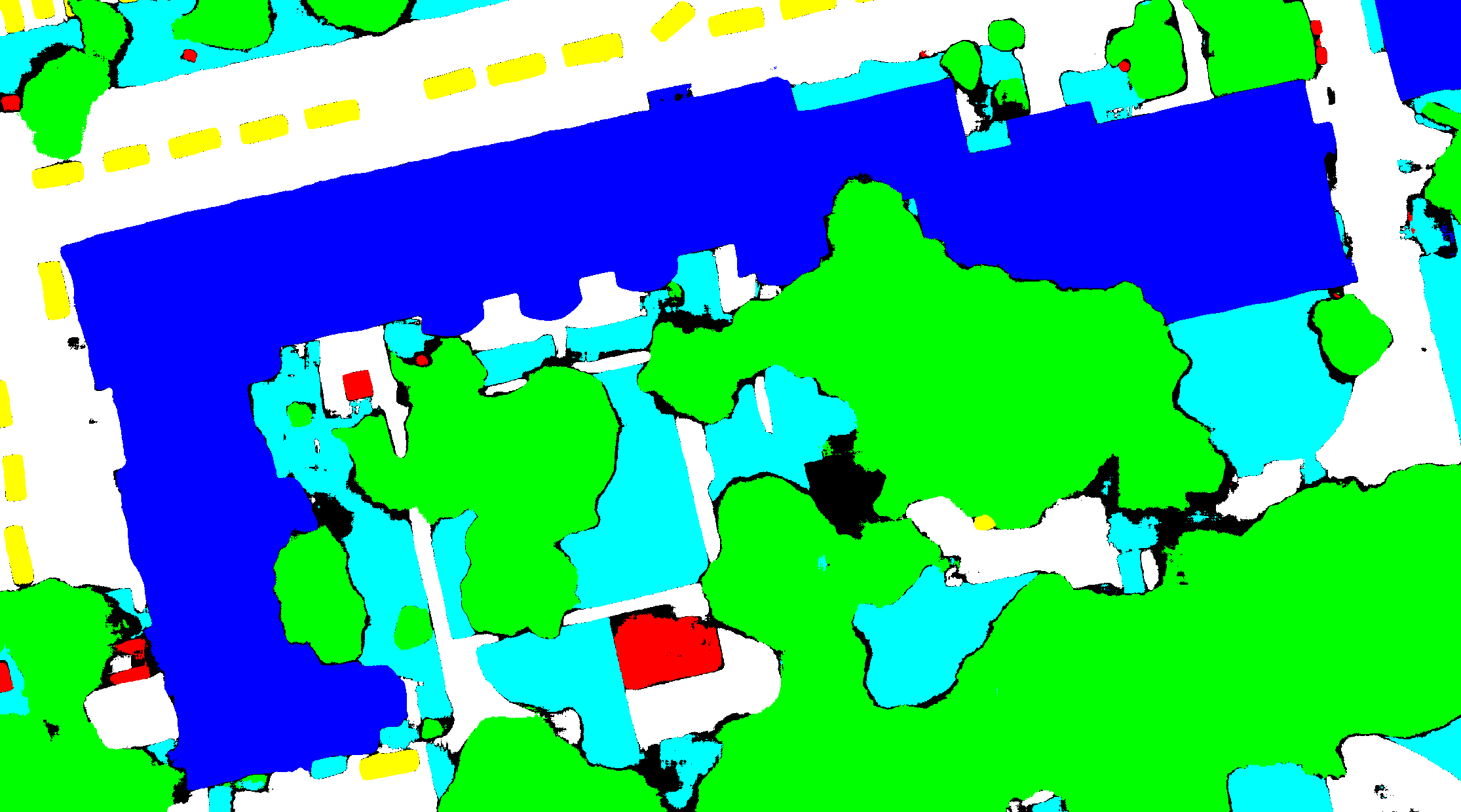}}\\[-0.2em]
\end{tabular}
\caption{Close-ups from the benchmarks. Classes: background (black), impervious surfaces (white), building (blue), low vegetation (cyan), high vegetation (green), car (yellow), and clutter (red). The results in the first two rows are from the Vaihingen dataset, and the last two rows are from the Potsdam benchmark.}
\label{fig:closeup_results_bechmarks}
\end{figure*}

We present the qualitative and quantitative experimental results on the benchmarks in a similar way described in Sec.~\ref{sec:experiments_on_luxcarta}. We report F1-score for each class in Tables \ref{table:vaihingen_f1} and \ref{table:potsdam_f1}, illustrate the plots of IoU vs. number of epochs on the Vaihingen benchmark in Fig.~\ref{fig:vaihingen_plots}, and show close-ups from both benchmarks in Fig.~\ref{fig:closeup_results_bechmarks}. As we use all the annotations at once for \textit{static learning}, we again choose this approach as the reference method.

From the plots in Figs. \ref{fig:luxcarta_plots} and \ref{fig:vaihingen_plots}, our first observation is that IoU values for each model, as the training iterations continue, fluctuate much more on the Luxcarta dataset than on the Vaihingen benchmark. We also observe that models, trained from the Vaihingen dataset converge faster. The reason for these two conclusions is that in the Vahingen dataset, a single aerial image was split into smaller tiles, while images in the Luxcarta dataset were taken from different cities at different dates; therefore, they have distinct color variations and visual features. Furthermore, the Luxcarta images are of much lower resolution, and the validation set consists of the cities that are not seen by the network during training. Because of all these reasons, accuracies for the same classes (i.e., \textit{building} and \textit{high vegetation}) are significantly lower in the experiments on the Luxcarta dataset than on the benchmarks.

Our observation for \textit{fixed representation} and \textit{fine-tuning} is similar to the experiments on the Luxcarta dataset. As can be seen in Fig.~\ref{fig:vaihingen_plots_fe}, for \textit{fixed representation}, although some classes such as \textit{impervious surface} and \textit{low vegetation} can be learned relatively well, the network performs poorly if the newly added class represents small objects like \textit{car}.

Since training as well as test tiles are from the same city, output of the memory network becomes almost the ground-truth for the previous classes. As a result, even if annotations for the previous classes are not accessible, new classes can be learned while exhibiting a similar performance for the former classes. We justify this claim in Fig.~\ref{fig:vaihingen_plots_inc}, in which it is demonstrated that IoU plots for the previously learned classes remain quite flat over time. The predicted maps of the close-ups from Vaihingen in Fig.~\ref{fig:closeup_results_bechmarks} for 3 approaches look very similar. The advantage of our approach is that with the help of the features for the previous classes, the network converges very fast for the new classes. For instance, as illustrated in Fig. \ref{fig:vaihingen_plots_joint}, it takes roughly 50 epochs in order for the network to converge for \textit{low vegetation} class when \textit{static learning} is applied, whereas with the proposed approach, a similar accuracy for the same class can be achieved in only a few epochs, as depicted in Fig.~\ref{fig:vaihingen_plots_inc}.

In this experimental setup, if the classes have distinct visual appearance and features like in the Vaihingen benchmark, as the classification tasks are shared between several classifiers, \textit{multiple learning} performs better especially when the class has a low number of samples such as \textit{car}. As training tiles of the Potsdam dataset contain the nDSM data, detecting \textit{car} class is easier on this dataset than on the Vaihingen benchmark. As reported in Table~\ref{table:potsdam_f1}, the gap between \textit{multiple learning} and the other approaches is smaller for this class. On the contrary, as can be seen in the last two rows in Fig.~\ref{fig:closeup_results_bechmarks}, \textit{clutter} class has high visual similarities with some pixels labeled as \textit{impervious surfaces} or \textit{low vegetation}. Hence, a single classifier that is trained jointly for all the classes, performs better in distinguishing these classes. Unlike \textit{multiple learning}, where several isolated classifiers are trained, our approach allows joint training via the memory network. Therefore, our approach performs better for these classes, as confirmed by Table~\ref{table:potsdam_f1}. The last row in Fig.~\ref{fig:closeup_results_bechmarks} exemplifies some mis-classified \textit{clutter} pixels by \textit{multiple learning} but correctly detected by our approach.

\section{Concluding Remarks}\label{sec:concluding_remarks}

We proposed a novel incremental learning methodology, which enables the neural network to learn segmentation capabilities for the new classes while retaining dense labeling abilities for the formerly trained classes without using the entire previous training data.

In our experiments, we first showed that the common learning approaches are extremely inefficient or inapplicable to learn from streaming data. We then demonstrated why using only the features extracted for the previous classes is inefficient to learn new classes. We also provided the results, showing that when the network is trained using only the new data without having any regularization, the learned information for the previous classes is catastrophically forgotten. Finally, on three different datasets we proved that the proposed approach achieves a high performance for the new classes without forgetting the old classes.

As the future work, we plan to explore how to incorporate domain adaptation techniques to our incremental learning methodology so that the trained network could better generalize to the data collected from new geographic locations.

\section*{Acknowledgment}
The authors thank ACRI-ST and CNES for initializing and funding this study. The authors also would like to thank Luxcarta\footnote{https://luxcarta.com} for providing the satellite images and their annotations.

\bibliographystyle{IEEEtran}
\bibliography{refs}

\begin{IEEEbiography}[{\includegraphics[width=1in,
height=1.25in,clip,keepaspectratio]{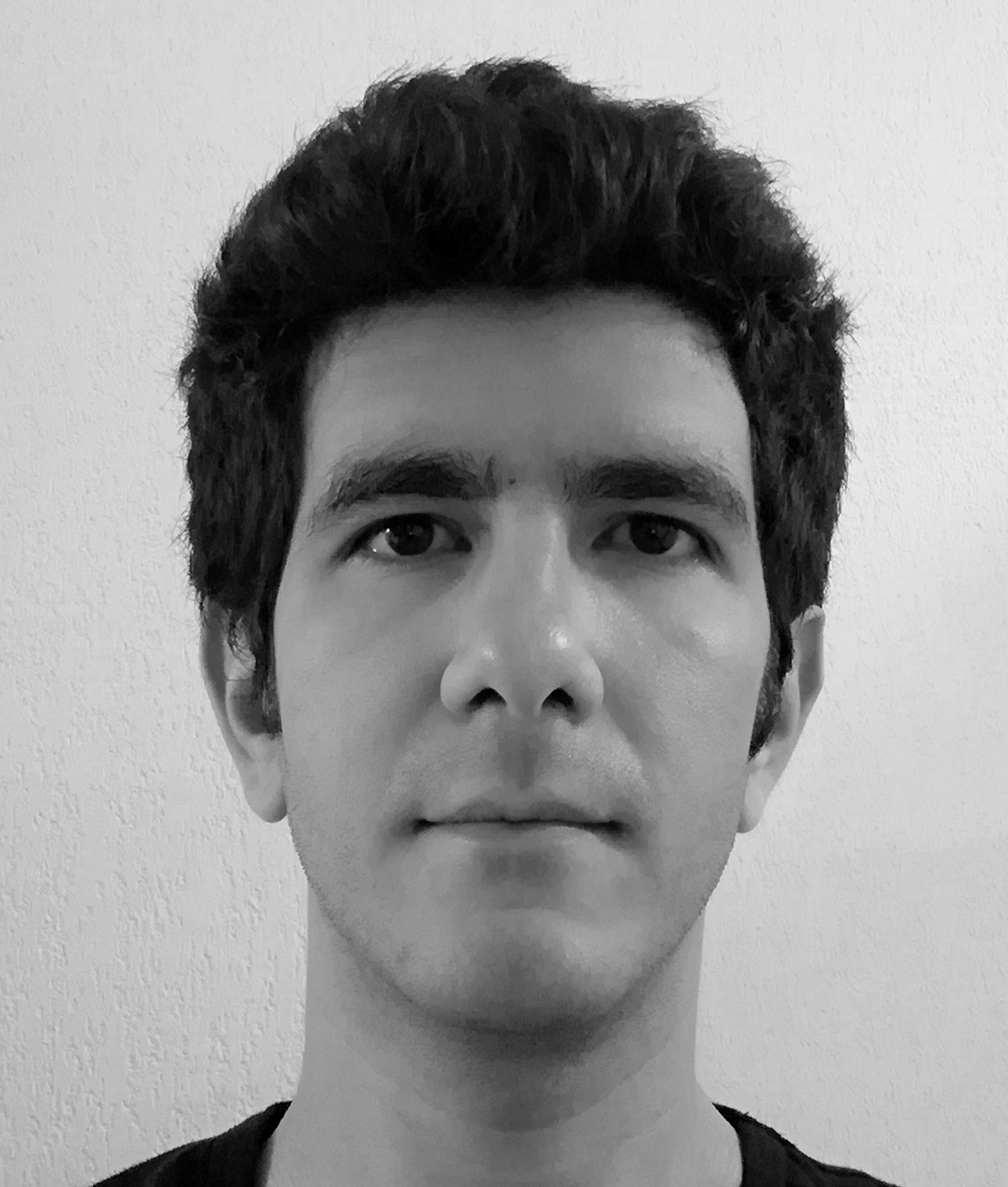}}]{Onur Tasar}
received the B.S. degree in computer engineering department from Hacettepe University, Ankara, Turkey in 2014, and the M.S. degree in computer engineering department from Bilkent University, Ankara, Turkey in 2017. He is currently working towards his Ph.D. at Inria Sophia Antipolis-M\'editerran\'ee within TITANE team, Valbonne, France.

His research interests include computer vision, machine learning, and deep learning with applications to remote sensing.
\end{IEEEbiography}

\vspace{-3mm}

\begin{IEEEbiography}[{\includegraphics[width=1in,
height=1.25in,clip,keepaspectratio]{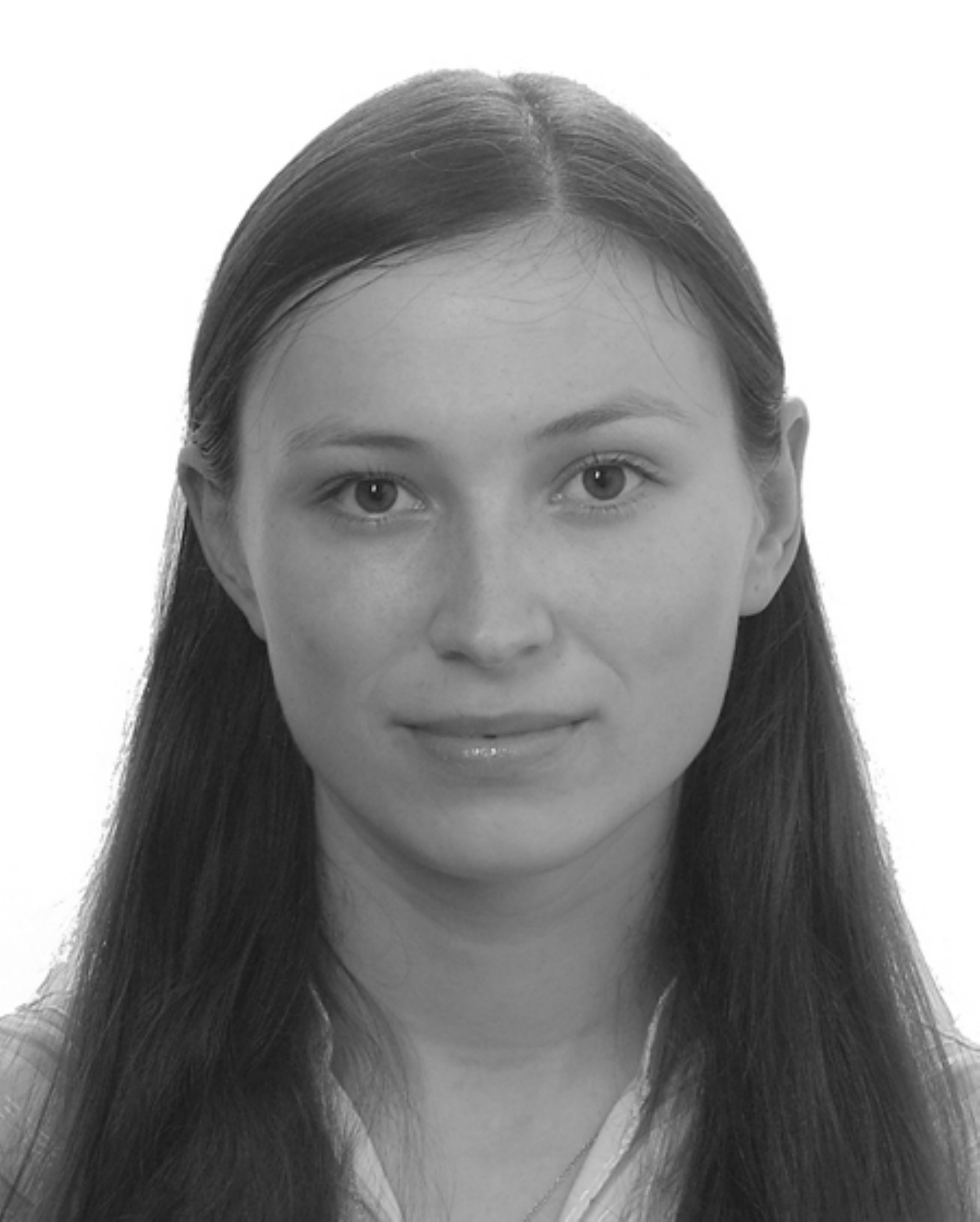}}]{Yuliya Tarabalka}

(S'08--M'10) received the B.S. degree in computer science from Ternopil Ivan Pul'uj State Technical University, Ukraine, in 2005 and the M.Sc. degree in signal and image processing from the Grenoble Institute of Technology (INPG), France, in 2007. She received a joint Ph.D. degree in signal and image processing from INPG and in electrical engineering from the University of Iceland, in 2010.

From July 2007 to January 2008, she was a researcher with the Norwegian Defence Research Establishment, Norway. From September 2010 to December 2011, she was a postdoctoral research fellow with the Computational and Information Sciences and Technology Office, NASA Goddard Space Flight Center, Greenbelt, MD. From January to August 2012 she was a postdoctoral research fellow with the French Space Agency (CNES) and Inria Sophia Antipolis-M\'editerran\'ee, France. She is currently a researcher with the TITANE team of Inria Sophia Antipolis-M\'editerran\'ee. Her research interests are in the areas of image processing, pattern recognition and development of efficient algorithms. She is Member of the IEEE Society.
\end{IEEEbiography} 

\vspace{-3mm}

\begin{IEEEbiography}[{\includegraphics[width=1in,height=1.25in,clip,keepaspectratio]{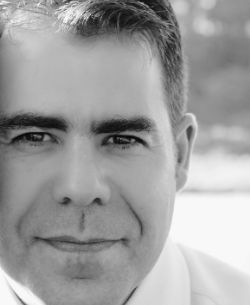}}]{Pierre Alliez}
Pierre Alliez is Senior Researcher and team leader at Inria Sophia-Antipolis - Mediterranee. He has authored scientific publications and several book chapters on mesh compression, surface reconstruction, mesh generation, surface remeshing and mesh parameterization. He was awarded in 2005 the EUROGRAPHICS young researcher award for his contributions to computer graphics and geometry processing. He was co-chair of the Symposium on Geometry Processing in 2008, of Pacific Graphics in 2010 and Geometric Modeling and Processing 2014. He was awarded in 2011 a Starting Grant from the European Research Council on Robust Geometry Processing. He is an associate editor of the ACM Transactions on Graphics.
\end{IEEEbiography}

\end{document}